\documentclass{article}
\usepackage[utf8]{inputenc}
\usepackage{graphicx} 
\usepackage{xcolor}
\usepackage{comment}
\usepackage{amsmath}
\usepackage{amsfonts}
\usepackage[preprint]{corl_2024}
\usepackage{float}
\usepackage{enumitem}
\usepackage[capitalise]{cleveref}
\usepackage{tikz}
\usepackage{siunitx}
\usepackage{listings}
\usepackage{subcaption}
\usepackage[export]{adjustbox}
\usetikzlibrary{shapes.geometric, arrows, positioning, calc, fit, backgrounds, shadows}

\definecolor{codegreen}{rgb}{0,0.6,0}
\definecolor{codegray}{rgb}{0.5,0.5,0.5}
\definecolor{codepurple}{rgb}{0.58,0,0.82}
\definecolor{backcolour}{rgb}{0.95,0.95,0.92}
\lstdefinestyle{mystyle}{
    backgroundcolor=\color{backcolour},   
    commentstyle=\color{codegreen},
    keywordstyle=\color{magenta},
    numberstyle=\tiny\color{codegray},
    stringstyle=\color{codepurple},
    basicstyle=\footnotesize,
    breakatwhitespace=false,         
    breaklines=true,                 
    captionpos=b,                    
    keepspaces=true,                 
    numbers=left,                    
    numbersep=5pt,                  
    showspaces=false,                
    showstringspaces=false,
    showtabs=false,                  
    tabsize=2
}
 
\DeclareMathOperator*{\mean}{mean}
\DeclareMathOperator*{\argmax}{argmax}

\author{
  Steven Morad\\
  Dept. of Computer Science and Technology\\
  University of Cambridge \\
  United Kingdom\\
  \texttt{sm2558@cst.cam.ac.uk} \\
  \And
  Ajay Shankar\\
  Dept. of Computer Science and Technology\\
  University of Cambridge \\
  United Kingdom\\
  \texttt{as3233@cst.cam.ac.uk} \\
  \And
  Jan Blumenkamp\\
  Dept. of Computer Science and Technology\\
  University of Cambridge \\
  United Kingdom\\
  \texttt{jb2270@cst.cam.ac.uk} \\
  \And
  Amanda Prorok\\
  Dept. of Computer Science and Technology\\
  University of Cambridge \\
  United Kingdom\\
  \texttt{asp45@cst.cam.ac.uk}
}
\date{April 2024}
\title{Language-Conditioned Offline RL for Multi-Robot Navigation}

\begin{document}
\maketitle
\begin{abstract}
    We present a method for developing navigation policies for multi-robot teams that interpret and follow natural language instructions. We condition these policies on embeddings from pretrained Large Language Models (LLMs), and train them via offline reinforcement learning with as little as 20 minutes of randomly-collected data. Experiments on a team of five real robots show that these policies generalize well to unseen commands, indicating an understanding of the LLM latent space. Our method requires no simulators or environment models, and produces low-latency control policies that can be deployed directly to real robots without finetuning. We provide videos of our experiments at \url{https://sites.google.com/view/llm-marl}.
\end{abstract}

\keywords{Multi-Robot, Multi-Agent Reinforcement Learning, Large Language Models}

\section{Introduction}
Natural language provides a rich and intuitive interface to describe robot tasks. For instance, commands such as ``navigate to the left corner" or ``pick up the can" lend themselves as more powerful and flexible alternatives to specifying $(x,y)$ coordinates or joint configurations. Using language descriptions to specify outcomes, particularly when interfacing with a team of robots, is thus a more natural choice, and one that does not require specially-trained operators.

Recent work on commanding robots with natural language tend to utilize large pretrained transformers \citep{vaswani_attention_2017} known as LLMs \citep{brown_language_2020, touvron_llama_2023, jiang_mistral_2023} or Large Multimodal Models (LMMs) \citep{openai_gpt-4_2024} for both language processing and control.
Often, the transformer receives a task and observation, and produces either an action or a sequence of actions to complete the task \citep{driess_palm-e_2023,dorbala_can_2024,brohan_rt-2_2023,shah_lm-nav_2023}.
The latter case reduces to open-loop control, which cannot adapt to uncertainty, while the former is limited by the high latency of these models, typically measured in seconds or hundreds of milliseconds, precluding them from dynamic scenarios.
These drawbacks are especially restricting in \textit{multi-agent} scenarios, where agents need closed-loop control to react and replan quickly based on the observed behavior of other robots~\citep{ni_bioinspired_2011,chen_decentralized_2017,blumenkamp_framework_2022}.
Developing multi-agent policies that can capitalize on the power of LLMs while retaining fast, reactive, and emergent team behavior is therefore a hard problem.

In this work, we present a method that maps high-level natural language commands directly to control outputs for a multi-robot system. First, we project natural language instructions into a latent space using a pretrained LLM, and then condition control policies on these latent embeddings. By factoring the LLM outside of the control loop, we achieve low-latency ($<\SI{2}{ms}$) realtime control. Our approach then generates a large multi-agent dataset using random real-world actions of a single robot. We train our task-conditioned policies on this real-world dataset using \emph{offline} Reinforcement Learning (RL). Because we train our policies on real data, we can deploy our policies back into the real world without any finetuning.

\paragraph{Contributions}
We claim the following key contributions:
\begin{itemize}[noitemsep,nolistsep,leftmargin=8pt]
    \item an architecture that enables low-latency multi-agent control through natural language;
    \item a method to generate large amounts of multi-agent training data from a single robot;
    \item findings that a one-line change to Q-learning improves offline training stability;
    \item evidence that our policies can generalize to unseen commands, solely through value estimation; and,
    \item the first demonstration of offline multi-agent RL on robots in the real-world.
\end{itemize}




\begin{figure}[t]
    \centering
    \includegraphics[trim={0cm 0cm 0cm 0cm},clip,width=0.32\linewidth]
    {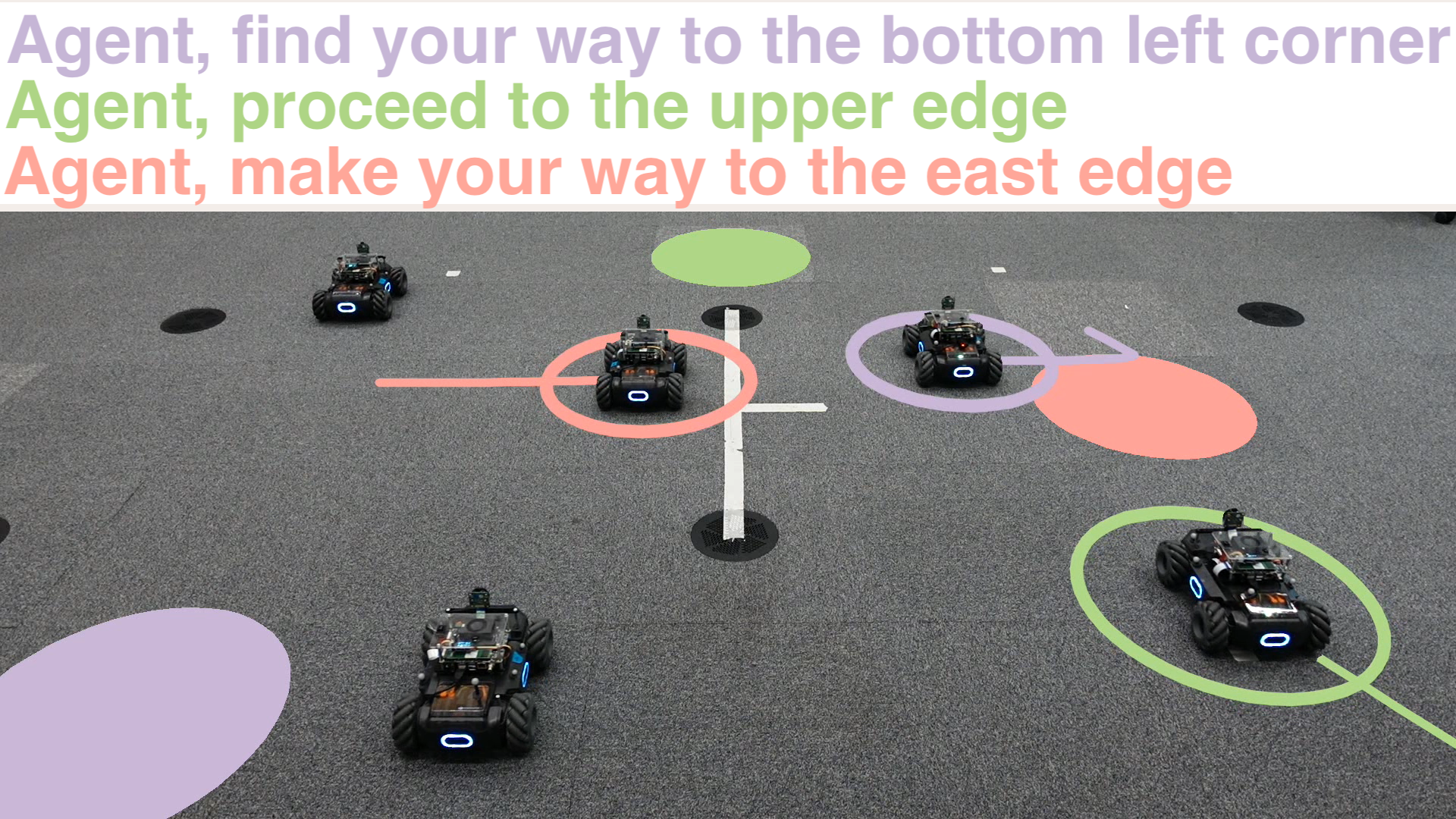}
    \includegraphics[trim={0cm 0cm 0cm 0cm},clip,width=0.32\linewidth]{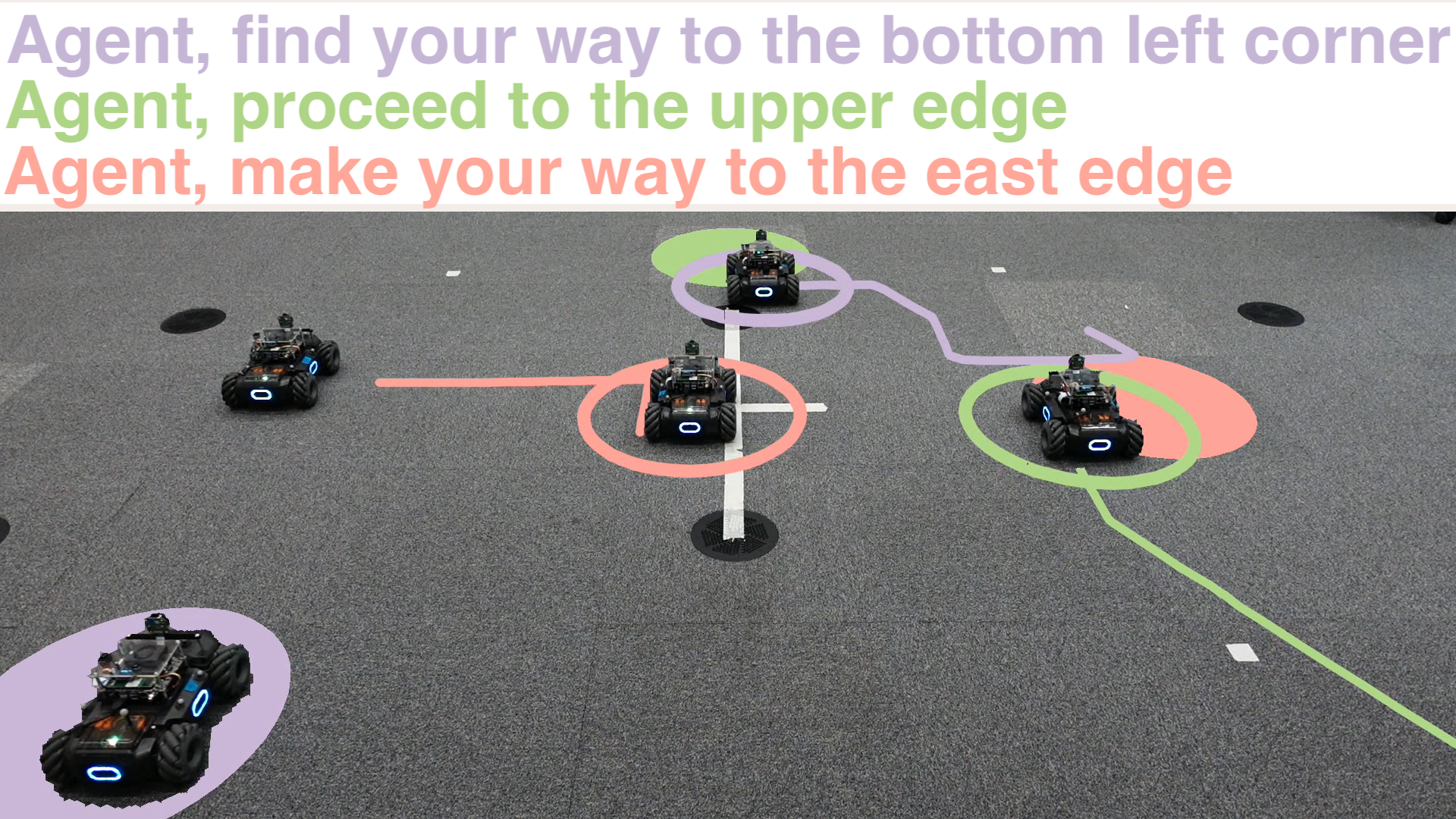}
    \includegraphics[trim={0cm 0cm 0cm 0cm},clip,width=0.32\linewidth]{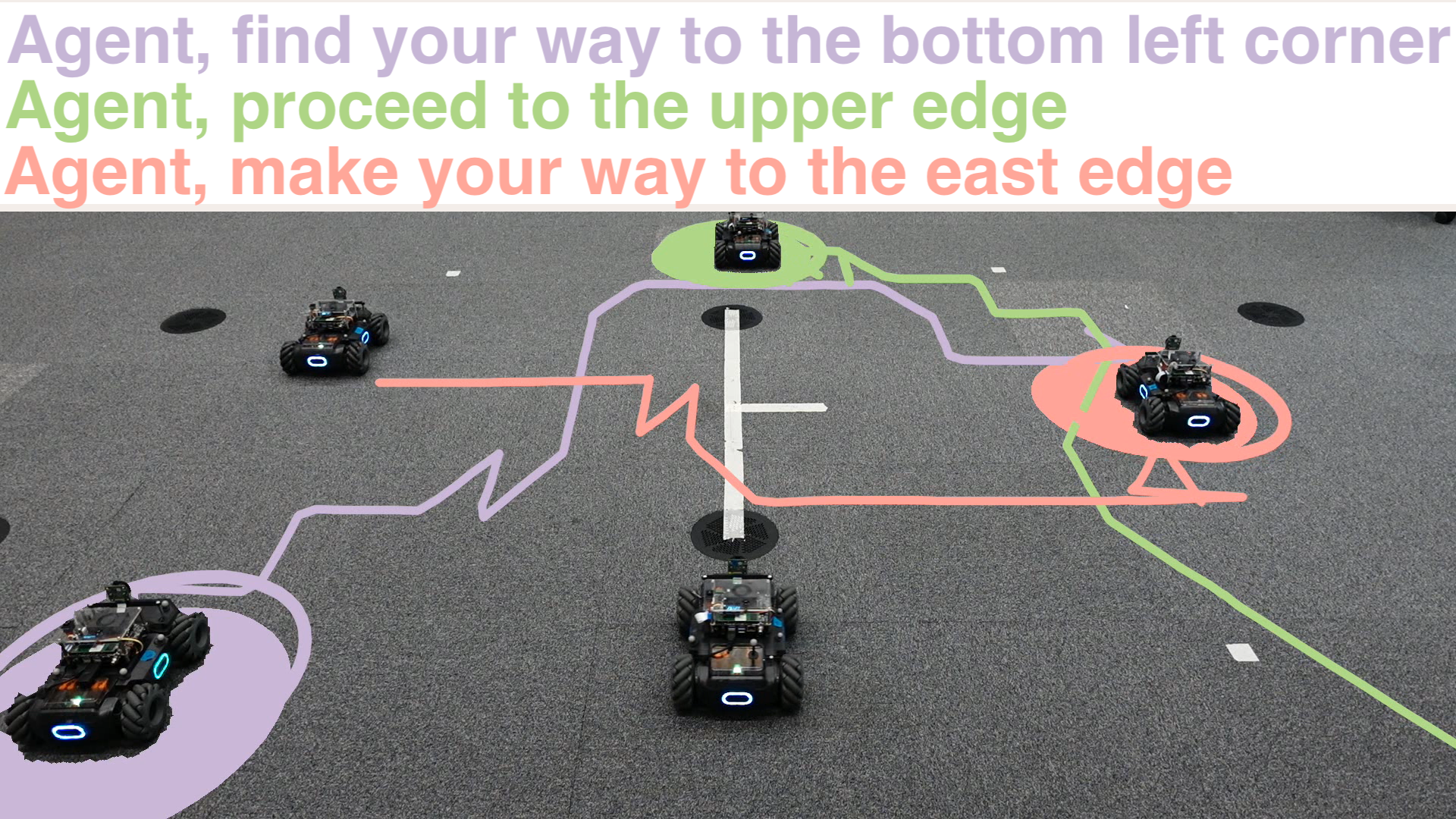}
    \caption{Our learned policies demonstrate emergent path deconfliction while following natural-language tasks. Agents, tasks, and goals are color-coordinated. Each agent (hollow disk) receives a natural language task (top), and must navigate to the goal (filled disk, rendered for visualisation only), leaving behind a colored trail. All five agents are in motion, but we highlight tasks and goals for three they execute a three-way yield. (Left) The red, purple, and green agents block each other from reaching their respective goals. (Middle) Green moves towards its goal and purple yields to green by moving north, and red yields to purple by remaining stationary. (Right) After green completes its task, purple and red complete their tasks.}
    \label{fig:demo}
\end{figure}

\section{Related Work}
LMMs such as GPT \cite{brown_language_2020} and LLMs such as LLaMa or Mistral \cite{touvron_llama_2023,jiang_mistral_2023} demonstrate strong emergent reasoning capabilities, mapping a series of input tokens to output tokens using a transformer architecture \citep{vaswani_attention_2017}. Given their similarities, we refer to both LLMs and LMMs as LLMs moving forward. LLMs often output tokens in the form of text, however, recent work has integrated them into robotic applications due to their powerful reasoning capabilities. For instance, PaLM-E \citep{driess_palm-e_2023} ingests image and text tokens and outputs text describing a list of steps to follow. Prior work demonstrates zero-shot navigation to visual targets by outputting text tokens that map to physical actions \cite{dorbala_can_2024}. In \cite{shah_lm-nav_2023}, the authors chain an LLM to a visual foundation model for visual navigation. Going one step further, the method in \cite{brohan_rt-2_2023} outputs tokens that correspond to actions, rather than text tokens. Certain works learn policies using imitation learning \citep{lynch_interactive_2023} or value functions \citep{ichter_as_2023} for textual action primitives output by an LLM. LLMs are often slow to execute, and these works are often not explicit about the latency between perception and action. For example, the smallest LLaMa-v2 model has roughly 100x fewer parameters than PalM-E model used in \cite{driess_palm-e_2023}. After 2-bit quantization and other efficiency tricks, LLaMa-v2 produces less than two tokens per second on an NVidia Jetson AGX Orin \citep{dfrobot_nvidia_2023}. Given that outputs often consist of multiple tokens, today's hardware precludes them from low-latency and reactive control tasks necessary in multi-robot systems.




The use of LLMs in multi-robot systems is still in its infancy, and similar to single-robot settings, relies on mapping text-tokens to plans or actions. Thus, they inherit the latency bottleneck of LLMs, compounded by generating tokens for an increased number of robots. Most prior work operates in simulation, with few papers demonstrating real-world implementations \citep{kannan_smart-llm_2024}. 
In \cite{mandi_roco_2023,chan_chateval_2023,zhang_building_2023}, individual agents converse in simulation to form a multi-agent system, with the agents eventually outputting text tokens for control.
LLMs resolve deadlocks in a simulated multi-robot navigation task, by outputting text tokens that correspond to agent directions \cite{garg_large_2024}.
Another approach allocates tasks to agents using an LLM \cite{kannan_smart-llm_2024}.
Like with single-agent LLM control, these approaches are generally not suitable for realtime control.

\section{Background}
\label{sec:rl}
\paragraph{Sentence Similarity LLMs} We are interested in a class of LLMs known as sentence similarity or feature extraction LLMs \cite{reimers_sentence-bert_2019,li_towards_2023,lee_nv-embed_2024}. An LLM maps $n$ input tokens map to $n$ latent representations at each transformer layer. The final layer predicts a token given $n$ latent representations. Rather than mapping to a token index, feature extraction LLMs pool over the final latent representation to return an embedding that summarizes the input text. These models are trained to have meaningful latent representations, such as semantic similarity corresponding to cosine similarity or Euclidean distance in the latent space. For example, the latent representation of ``quick brown fox'' would be closer to ``fast maroon badger'' than ``machine learning''. 

\paragraph{Reinforcement Learning}
We refer the reader to \cite{sutton_reinforcement_2018} for a proper treatment of reinforcement learning. Reinforcement learning aims to learn a policy $\pi$ to maximize cumulative rewards in a Markov Decision Process (MDP) $\langle S, A, R, T, \gamma \rangle$. The agent selects actions $a \sim \pi(s)$ to transition to new states $T(s' | s, a)$ and receive rewards $r = R(s, a, s')$, with the goal of maximizing
$
\mathbb{E}_{\pi} \left[ \sum_{t=0}^\infty \gamma^t R(s, a, s') \right].
$
For value-based policies, we will often attempt to learn the value of states $s$ using either a value function 
$
    V(s_0) = \mathbb{E}_{\pi} \left[ \sum_{t=0}^\infty \gamma^t R(s, a, s') \right]
$, or a Q function 
$
    Q(s_0, a_0) = R(s_0, a_0, s_1) + 
 \mathbb{E}_{\pi} \left[ \sum_{t=1}^\infty \gamma^t R(s, a, s') \right].
$
We then follow a policy that maximizes $V$ or $Q$, such as the greedy policy
$
    \pi(s) = \argmax_{a \in A} Q(s, a).
$

\paragraph{Multi-Agent RL}
We direct the reader to \citep{yang_overview_2021} for a review of Multi-Agent Reinforcement Learning (MARL).
It has a number of formulations, depending on the structure of the reward functions.
We extend the MDP above to a Decentralized MDP with factorized rewards, defined as $\langle S, \mathcal{A}, T, \mathcal{R}, \mathcal{O}, \Omega, \gamma \rangle$. Each agent $i$ has its own action space, reward function, and observation function: $\mathcal{A} = \{A_i\}_{i=1}^n, \mathcal{R} = \{R_i\}_{i=1}^n, \mathcal{O} = \{O_i\}_{i=1}^n$. The state $s \in S$ is a function of local observations $s = f(o_1, \dots, o_n)$, where each observation $o_i \in \Omega$ is found via $o_i \sim O_i(s)$. The state transition function is $T: S \times \mathcal{A} \mapsto S$, and the reward function $R_i(s, a, s')$ is defined on the states. There are a variety of MARL algorithms that focus on multi-agent credit assignment, such as MADDPG \citep{lowe_multi-agent_2017}, QMix \citep{rashid_monotonic_2020}, and COMA \citep{foerster_counterfactual_2018}. Given our factorized reward function, we choose to learn independent policies using algorithms like Independent Proximal Policy Optimization (IPPO) \citep{de_witt_is_2020} or Independent Q-Learning (IQL) \citep{tan_multi-agent_1993}. Additionally, \citep{pan_regularized_2021} proposes a softmax-regularized form of the Q-learning objective that provides better results in MARL.

\paragraph{Offline RL}
Offline RL learns a policy from existing data, which is especially useful in robotics, where simulators can be inaccurate and data collection can be costly. Unlike imitation learning which requires expert data, offline RL can learn policies from \emph{completely random data}.
In offline RL, we no longer have direct access to the MDP or DecMDP.
Instead, a behavior policy $\pi_\beta$ collects transition tuples $\langle s, a, r, s' \rangle$ (or $\langle o_i, a_i, r_i, o'_i \rangle$ in the multi-agent case).
With fixed datasets, standard value functions tend to \emph{overextrapolate} -- incorrectly estimating values for out-of-distribution transitions, which propagate via bootstrapping. Solutions like Conservative Q-Learning (CQL) regularize value estimates to prevent positive overextrapolation \cite{kumar_conservative_2020, kumar_stabilizing_2019, shao_counterfactual_2023}, while others constrain the learned policy to be close to the data collection policy $\pi_\beta$ \cite{fujimoto_minimalist_2021, ball_efficient_2023}. There is little work combining MARL with offline RL, likely because both offline RL and MARL are difficult to solve individually. Prior work proposes counterfactual offline multi-agent Q-learning based on CQL \cite{shao_counterfactual_2023,kumar_conservative_2020} and Implicit Constraint Q-Learning (ICQ), which uses soft Q-learning to prevent overextrapolation \cite{yang_believe_2021}. 

\paragraph{Task-Conditioned Policies}
There are a number of names for task-conditioned reinforcement learning. It is sometimes called goal-conditioned RL, multi-task RL, or universal value function approximation \citep{liu_goal-conditioned_2022,teh_distral_2017,schaul_universal_2015}.
Although formulated in different ways, the key difference between these and `standard' RL is the introduction of a \emph{task} or \emph{goal} $g \in G$ into the reward and value functions, as well as the policy:
$
    \langle R(s, a, s'), V(s), Q(s, a), \pi(s) \rangle$ becomes    
$
\langle R(s, a, s' \vert g), V(s \vert g), Q(s, a \vert g), \pi(s \vert g) \rangle.
$
Thus, instead of learning a value function and policy for a specific task, we learn one that can generalize over a task space $G$.





\section{Approach}
Our objective is to train multi-agent policies to follow natural language navigation tasks. Our approach consists of two distinct parts, \textit{creating the dataset} and \emph{training the model}. To generate the dataset, we first collect real-world data using a \emph{single robot}. Then, we generate a large quantity of natural language tasks. We sample real-world data and tasks independently, combining them to form a large multi-agent dataset. Finally, we train a policy on the multi-agent dataset using offline RL.


\subsection{Dataset Creation and Augmentation}
\label{sec:dataset}
We use the DJI RoboMaster, a four-wheel mecanum-drive holonomic robot, for our experiments \citep{blumenkamp_cambridge_2024}.
We collect a dataset $\mathcal{D}$ by recording $\langle s, a, s' \rangle$ tuples from a single robot as it uniformly samples actions from the action space ($a \sim \mathcal{U}(A)$).
We record one tuple per second, resulting in a total of 5400 tuples spanning 90 minutes of data collection.
The state space $S$ consists of two-dimensional position and velocity vectors, while the action space $A$ consists of 9 discrete actions, corresponding to eight direction vectors (at \SI{0.3}{m/s} each), plus an action that corresponds to zero velocity.

\paragraph{Generating Task Embeddings}
Each task $g \in G$ is a natural language task, such as \texttt{Agent, the south west corner is your target}.
We encode each task into a latent embedding $z = \phi(g)$ using a feature extraction LLM. Each agent in the multi-agent system independently receives tasks, and the tasks differ between agents. For example, one robot might receive the task \texttt{Agent, the south west corner is your target}, while another might receive \texttt{Agent, navigate to the left edge}. We split our tasks and embeddings into a set of \emph{train tasks} and \emph{test tasks}. We use 396 tasks for the train set and eight tasks for the test set (\cref{sec:appendix_tasks}). Each item in our test set is a \emph{never before seen} command, enabling us to gauge our model's generalization capabilities.

\paragraph{Combinatorial Multi-Agent Augmentation}
Rather than collect a multi-agent dataset, we choose to collect a single agent dataset because it can be exploited to create a large multi-agent dataset in certain scenarios. To turn our single agent dataset into a multi-agent one, we combine single agent transitions into a multi-agent transition
\begin{align}
    s = \langle s_i, s_j, s_k, \dots \rangle, \,
    a = \langle a_i, a_j, a_k, \dots \rangle, \,
    s' = \langle s'_i, s'_j, s'_k, \dots \rangle, \quad
    {i, j, k} \in \mathcal{U}(|\mathcal{D}|).
\end{align}
This process creates a large amount of multi-agent data from a small amount of single agent data. For example, from our 5400 single agent transitions, we can create a three agent dataset with $\textrm{Permute}(5400, 5) \approx 10^{18}$ transitions.
Given our 396 task embeddings and five agents, we can generate $\textrm{Permute}(396, 5) \approx 10^{12}$ multi-agent task configurations. The product of these two sets results in $10^{30}$ possible task-transition configurations. Given our sample rate, this would equate to collecting data for a duration $10^{12}$ times the age of the universe -- for this reason, we never physically materialize the multi-agent dataset. Rather, we construct these multi-agent transitions on the fly by uniformly sampling transitions with replacement $\langle s_i, a_i, s_i' \rangle \sim \mathcal{U}(\mathcal{D})$ and tasks without replacement $g_j \sim \mathcal{U}(G)$. 

\paragraph{Reward and Termination Functions}
For each agent $i$, we create a task-conditioned reward function $R_i$ and episode termination function $D_i$. We compute the reward and termination conditions across various tasks $g_j$ and corresponding goal coordinates $r_j$ as
\begin{align}
    R_i(s, a, s', g_j) &= (\lVert (p'_i - r_j) \rVert_2 -  \lVert (p_i - r_j) \rVert_2) (1 + v'_i + v_i) - c(p'_i, s) \\ D_i(s, a, s', g_j) &= c(p'_i, s),
\end{align}
where $p_i, v_i$ corresponds to the position and velocity portions of the state for agent $i$. The function $c$ computes collisions between individual agents as well as with the walls. The first term in the reward function provides a reward for moving closer to the goal, scaled by a velocity to create smoother movement. The second reward term penalizes collisions. This reward function ensures that each agent reaches its goal quickly while avoiding collisions, similar to prior work \citep{bettini_vmas_2022}.

\subsection{Training Objectives}
\begin{figure}[t]
    \centering
    \includegraphics[width=0.8\linewidth]{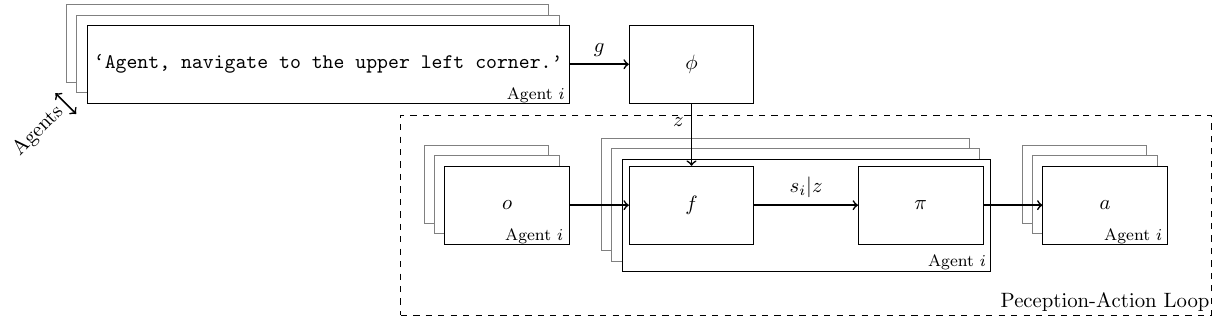}
    \caption{Our proposed multi-robot model architecture. Each agent receives a different natural language task and a local observation. We summarize each natural language task $g_i$ into a latent representation $z_i$, using an LLM $\phi$. The function $f$ is a graph neural network that encodes local observations $o_1, o_2, \dots $ and task embeddings $z_1, z_2, \dots $ into a task-dependent state representation $s_i | z$ for each agent $i$. We learn a local policy $\pi$ conditioned on the state-task representation. Functions $\pi, f$ are learned entirely from a fixed dataset using offline RL. Because we compute $z_i$ only once per task, the LLM is not part of the perception-action loop, allowing the policy to act quickly.}
    \label{fig:hero}
\end{figure}
\label{sec:training}
As stated in \cref{sec:rl}, there are a number of offline RL algorithms, however, the majority are designed for single agent setups. Furthermore, most focus on continuous action spaces and thus utilize KL divergence, causing a divide-by-zero error when considering the greedy policy. Instead, we reformulate Q-learning through the lens of \emph{Expected SARSA}, a relatively understudied \citep{van_seijen_theoretical_2009} form of value approximation from the Sutton and Barto textbook \cite{sutton_reinforcement_2018}. In the following subsection, we show how Expected SARSA subsumes approaches like Max Q-learning, softmax-regularized Q-learning \citep{pan_regularized_2021}, and Mean Q-learning \citep{emmons_rvs_2021}. Furthermore, we can implement Expected SARSA on top of Q-learning by changing a single line of code (\cref{sec:appendix_sarsa}).

\paragraph{Extending Expected SARSA to Offline Datasets}
Expected SARSA results in lower variance updates than standard Q-learning \citep[6.6]{sutton_reinforcement_2018}, which is useful in offline learning scenarios where a single bad update can trigger instabilities that result in runaway overextrapolation of Q values \citep[6.6]{sutton_reinforcement_2018, fujimoto_minimalist_2021}. Expected SARSA generally results in safer, albeit less optimal policies than Q-learning, as demonstrated in the famous cliffwalker experiment \cite[6.5]{sutton_reinforcement_2018, van_seijen_theoretical_2009} (\cref{sec:appendix_sarsa}). This is especially useful in robotics, where safer policies may be preferred over optimal policies. More formally, Expected SARSA learns a policy-weighted value approximation 
\begin{align}
    Q_{\pi}(s, a) = R(s, a, s') + \gamma \sum_{a' \in A} \pi(a' | s') Q_{\pi}(s', a'). \label{eq:expected_sarsa}
\end{align}
However, Expected SARSA is not designed for offline RL. Normally, a stochastic exploratory policy is slowly annealed into a optimal deterministic Max Q policy. In offline RL, we cannot access the MDP and thus cannot explore. Therefore, it is unclear what to use for $\pi$ in \cref{eq:expected_sarsa}. If we plug in the standard greedy policy, Expected SARSA decomposes into the standard Max Q-learning objective
\begin{align}
    Q_*(s, a) &= R(s, a, s') + \gamma \sum_{a' \in A}  \delta^{a'}_{\argmax_{a' \in A}  Q_*(s', a')}  Q_*(s', a') \\ 
    &= R(s, a, s') + \gamma \max_{a' \in A} Q_*(s', a') \label{eq:maxq}
\end{align}
where $\delta$ is the Kronecker delta (indicator) function. As discussed in \cref{sec:rl}, Max Q-learning struggles in purely offline scenarios. On the other hand, using the collection policy $\pi = \pi_\beta$ in the objective could prevent overextrapolation at the expense of less optimal behavior. Given that our dataset was collected by sampling actions from a uniform distribution $\pi_\beta = \mathcal{U}(|A|)$, substituting $\pi_\beta$ into Expected SARSA results in a Mean Q-learning objective similar to \cite{emmons_rvs_2021}
\begin{align}
    Q_\beta (s, a) &= R(s, a, s') + \gamma \sum_{a' \in A} \mathcal{U}(|A|) Q_\beta(s', a') = R(s, a, s') + \gamma \sum_{a' \in A} \frac{1}{|A|} Q_\beta(s', a') \\
    &= R(s, a, s') + \gamma \, \mean_{a' \in A} Q_\beta(s', a'). \label{eq:meanq}
\end{align}
We can also parameterize the space of policies between  $\pi$ and $\pi_\beta$ using the softmax operator. This corresponds to optimizing the softmax-regularized MARL objective \citep{pan_regularized_2021}, and also resembles the ICQ objective \citep{yang_believe_2021}
\begin{align}
    Q_{\sigma}(s, a) = R(s, a, s') + \gamma \, \left[ \sum_{a' \in A} \left( \frac{e^{Q_{\sigma}(s', a') / \tau}}{\sum_{\alpha' \in A} e^{Q_{\sigma}(s', \alpha') / \tau}} \right) Q_{\sigma}(s', a') \right] \label{eq:softq}
\end{align}
where $\tau$ is a temperature hyperparameter. As $\tau \rightarrow 0$, Soft Q becomes Max Q, and as $\tau \rightarrow \infty$, Soft Q becomes Mean Q. We investigate all of these objectives in our experiments.

\paragraph{Model Architecture}
Given that we can factorize rewards in the navigation scenario, we opt for a variant of Expected SARSA in the style of IQL with shared observations \citep{tan_multi-agent_1993}. Each agent approximates their own embedding of the global state $s_1, s_2, \dots, s_m \in S$, given a set of local observations $o_1, o_2, \dots, o_n \in O$ using a function $f$ (\cref{fig:hero}). Although it is conceptually simpler to model $f: O^n \mapsto S, \, \pi(s_i, z_i)$, we find that conditioning the state $s_i$ on $z$ performs better in practice: $f: O^n \times Z^n \mapsto S, \, \pi (s_i | z)$. This formulation enables agents to tailor their state representation based on the assigned tasks. Given $n$ agents, we model $f$ using a graph neural network over a fully-connected graph. We compute each $s_i$ using a pairwise nonlinear function over each robot's task and sensor information
\begin{align}
    s_i = \frac{1}{n} \sum_{1 \leq j \leq n} B(o_i \vert \vert z_i \vert \vert o_j \vert \vert z_j), \quad B(x) = 
    \sigma[ \mathrm{LN}(Wx + b)],
\end{align}
where $\vert \vert$ represents concatenation $B$ and represents a block with a learnable parameters $W, b$, layer normalization LN \citep{ba_layer_2016}, and leaky ReLU activation $\sigma$ \citep{xu_empirical_2015}. Each agent predicts their Q value $Q(s_i | z, a_i)$ independently, following one of the previous Q-learning methods from \cref{sec:training}. Our Q function consists of two consecutive blocks $B$, followed by a dueling Q linear layer \citep{wang_dueling_2016}. We use the clipped double Q trick \citep{fujimoto_addressing_2018} to further increase training stability.

\section{Experiments and Discussion}
Our experiments aim to answer four key questions:
(1) Is it possible to generalize to an LLM latent space?
(2) Which loss function should we use to train our policy?
(3) How much data must we collect to train a suitable policy?
(4) How does our policy perform on real robots? 

\paragraph{LLM Latent Space Analysis}
In our first experiment, we quantify how likely a policy is to generalize to the LLM latent space. Recall that we map a natural language goal to a latent embedding using an LLM $z = \phi(g)$. Now, we attempt to learn a decoder $\psi$ that maps the latent goal embedding to an alternative representation of the goal $\overline{g} = \psi(z)$. If we can correctly predict $\overline{g}$ for a never-before-seen $g$, we say that the decoder has \emph{generalized} to the latent space. To this end, we train an MLP decoder to regress a 2D goal coordinate. For example, the embeddings of \texttt{`navigate to the left edge'} and \texttt{`the west edge is your target'} should both decode to $\overline{g} = (-k, 0)$ for some given $k$. If our learned decoder $\psi$, never having seen the phrases \texttt{`go to the'} or \texttt{`north east corner'}, can correctly decode \texttt{`go to the north east corner'} to $\overline{g} = (k, k)$, then we say that our decoder has generalized to the LLM latent space.

\begin{figure}
    \centering
    \begin{minipage}{0.7\linewidth}
    \vspace{0em}
    \includegraphics[width=\linewidth]{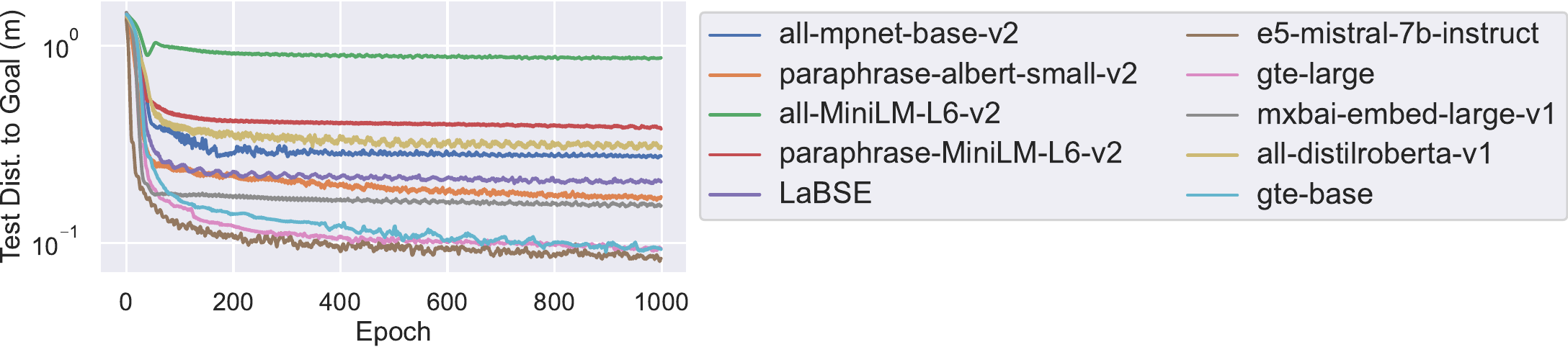}
    \end{minipage}\hfill
    \begin{minipage}{0.3\linewidth}
    \vspace{0em}\hfill
    \begin{tabular}{cc}
        Agents & Latency (ms) \\
        \hline
        1 & $0.57 \pm 0.14$  \\
        2 & $1.03 \pm 0.07$ \\
        3 & $1.13 \pm 0.05$ \\
        4 & $1.40 \pm 0.11$ \\
        5 & $1.51 \pm 0.10$
    \end{tabular}
    \end{minipage}
    \caption{(Left) A comparison of LLMs \citep{li_angle-optimized_2023, feng_language-agnostic_2022, muennighoff_mteb_2023, lee_open_2024, reimers_sentence-bert_2019,li_towards_2023,jiang_mistral_2023,wang_improving_2024} used for feature extraction. Our decoder only generalizes to certain LLM latent spaces (note the log scale y-axis). (Right) Control latency (observation to action) for different team sizes, tested on a 2020 MacBook Air CPU. Our policies map perception to action much faster than those with an LLM in the perception-action loop, which can often take seconds to produce each action.}
    \label{fig:llm-compare}
\end{figure}

We compare several LLMs and plot the mean position test error in \cref{fig:llm-compare}, utilizing the train and test split from \cref{sec:dataset}. We find that certain LLMs are not suitable for our purposes (e.g., MiniLM has an error of nearly \SI{1}{m}). The GTE-Base LLM \cite{li_towards_2023} is a good tradeoff between latent size and performance, with a mean error of \SI{7}{cm} across never-before-seen tasks. Therefore, we use the GTE-Base LLM in all further experiments. 

\subsection{Simulation Experiments}
Although our approach does not require a simulator to learn a policy, they can be helpful for analysis. To this end, we learn a single-agent state transition model from our dataset
$
   s_i' = T_i(s_i, a_i),
$
where $T_i$ is approximated using a deep neural network. We create a multi-robot simulator via $T(s, a) = \{ T_i(s_i, a_i)\}_{i=1}^n $. This model is not perfect -- for example, we can detect collisions between agents but cannot model them. That said, it still allows us to probe various objective functions.

\begin{figure}
    \centering
    \includegraphics[width=0.32\linewidth]{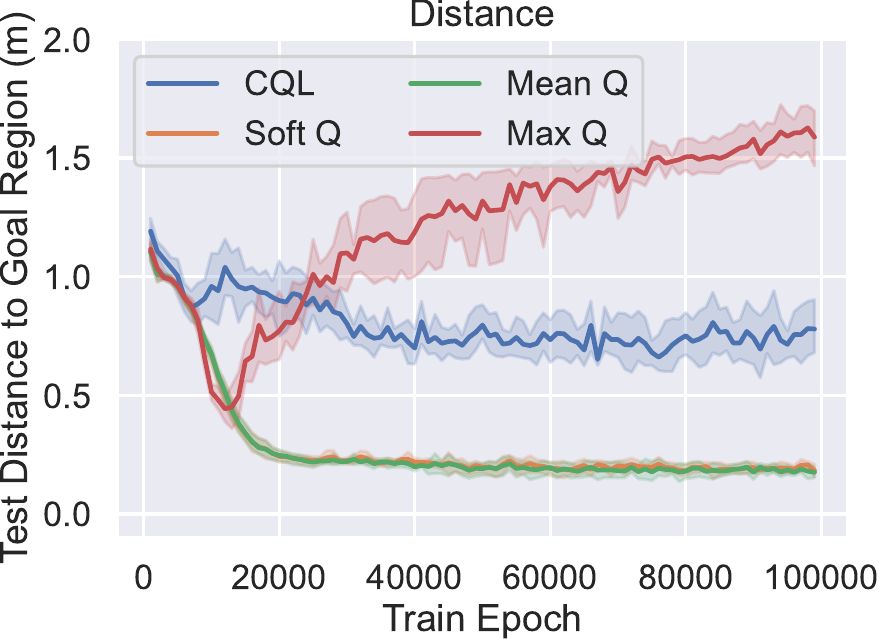}\hfill
    \includegraphics[width=0.32\linewidth]{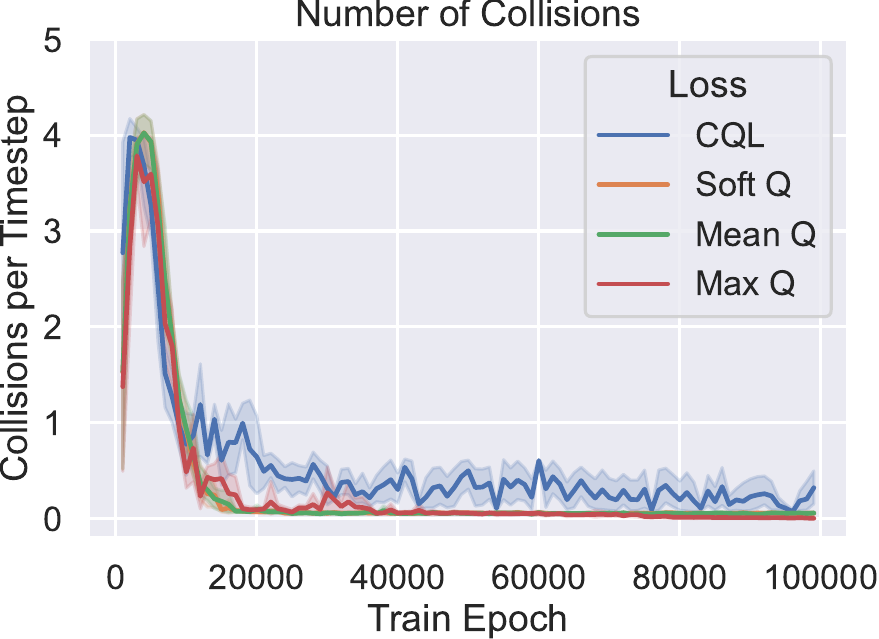}\hfill
    \includegraphics[width=0.32\linewidth]{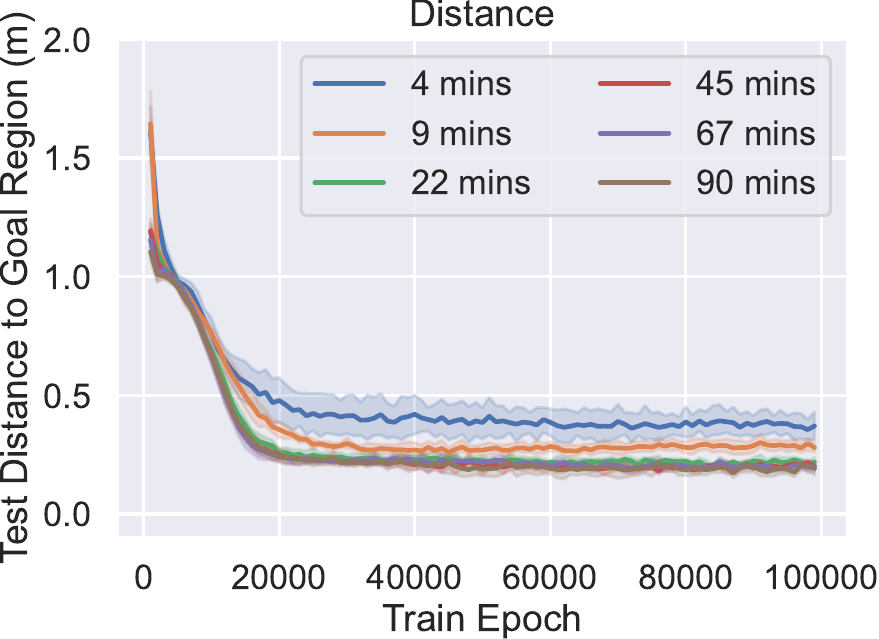}
    \caption{(Left Two) We compare the best CQL and Soft Q variants to Max Q and Mean Q objectives. Soft Q and Mean Q perform best. (Rightmost) We experiment with how much data collection is necessary to train a sufficient policy, using the Mean Q objective.}
    \label{fig:sim_results}
\end{figure}

\paragraph{Objective Function Analysis}
In \cref{sec:training}, we discuss a number of optimization objectives. Using our simulator, we compare the resulting policies trained using each of these objectives. We report metrics on the \emph{test tasks} which were not seen during training (\cref{sec:dataset}). In \cref{sec:reg_ablation}, we explore various CQL regularization strengths and Soft Q temperatures for an Expected SARSA Soft Q objective. We compare the best performing CQL and Soft Q variants to the Max Q and Mean Q weightings in \cref{fig:sim_results}, along with perception to action latency. We report the mean distance to the goal over an episode, as well as the number of collisions. 

Max Q begins to overextrapolate around 15k epochs, similar to findings in prior work \citep{fujimoto_minimalist_2021,levine_offline_2020}. CQL's regularization improves upon the Max Q performance. Both Mean Q and Soft Q perform comparably, outperforming both CQL and Max Q with respect to distance and number of collisions. The strong performance of Mean Q suggests that strong regularization towards the dataset collection policy $\pi_\beta$ is necessary, even though the regularization reduces optimality. This echos findings from \citep{emmons_rvs_2021}, where maximizing the Mean Q Monte Carlo return can outperform CQL. Both Mean Q and Soft Q objectives have nonzero weight over all state-action values. For every state, they always consider the in-distribution action collected during training, which can ``ground'' the value estimate and further prevent overextrapolation error.

\paragraph{Data Efficiency}
We investigate how the policy degrades as we reduce the amount of training data available. We subsample our dataset into smaller datasets, and report the results in \cref{fig:sim_results} right). We find the performance remains nearly the same with as little as 22 minutes of data. Our results also show that we can learn an imperfect yet reasonable policy from just four minutes of real world data. We suspect this is possible for two reasons: (1) Our nearly four hundred tasks correspond to nearly four hundred losses for each transition in our dataset. (2) The way we build our multi-agent dataset allows for a combinatorial number of possible agent configurations (\cref{sec:dataset}).

\subsection{Real World Experiments}
\begin{figure}
    \centering
    \includegraphics[width=0.32\linewidth]{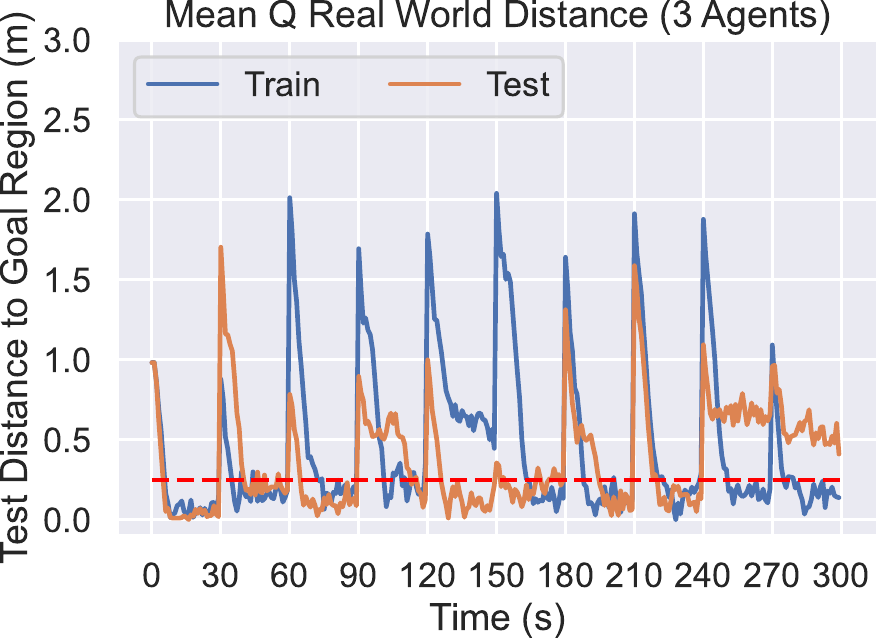}\hfill
    \includegraphics[width=0.32\linewidth]{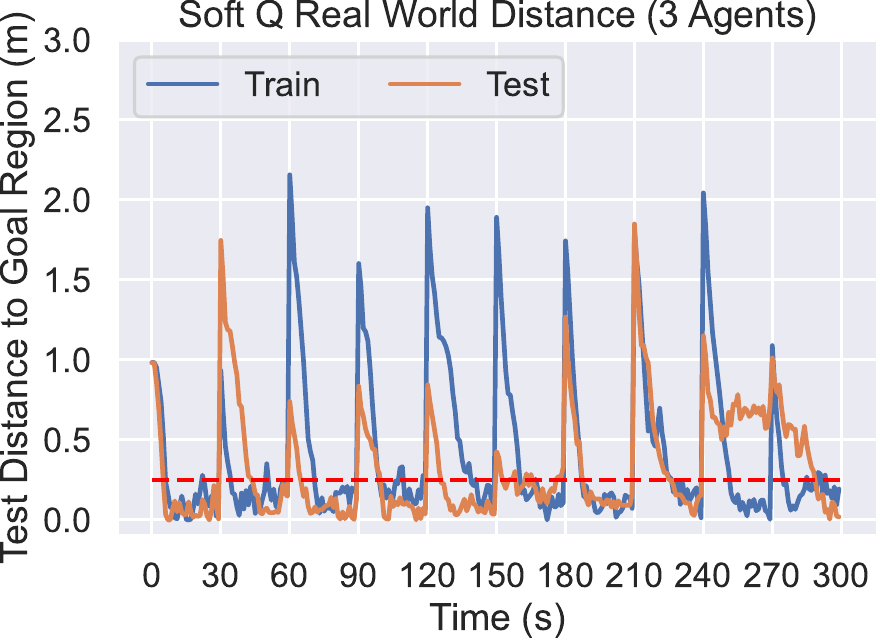}\hfill
    \includegraphics[width=0.32\linewidth]{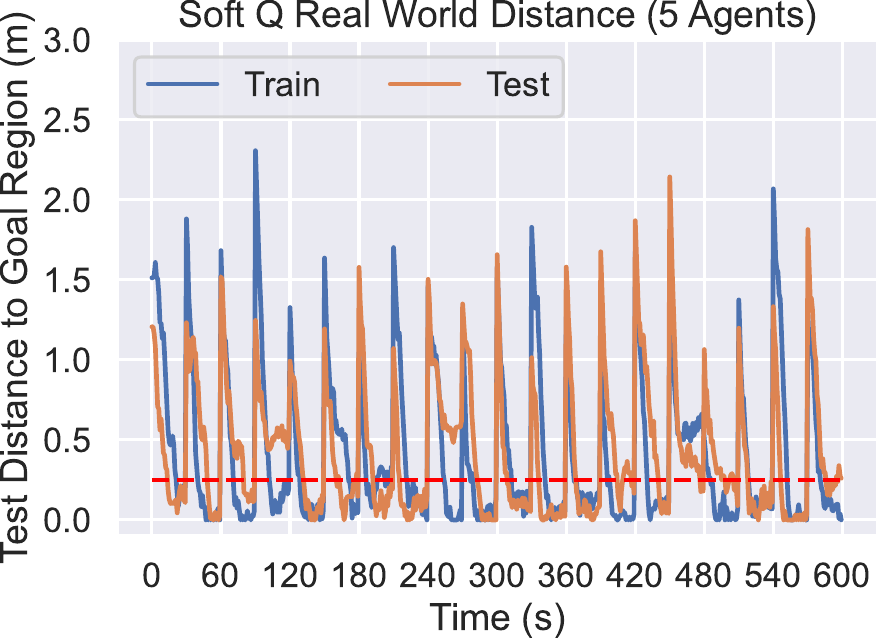}
    \caption{Evaluations from real-world multi-agent navigation tests, 
    where each agent is provided a new task every \SI{30}{s}. We plot their distance from the goal (averaged over agents) as they navigate. The blue line represents tasks the agent has seen before, while the orange line represents unseen tasks. Our results show that our agents are able to generalize to unseen, out-of-distribution tasks. We consider success as reaching a mean distance $< \SI{25}{cm}$ (red line) from the goal regions. For three agents, the Mean Q policy solves 9/10 train tasks and 8/10 test tasks while the Soft Q policy solves 10/10 train tasks and 9/10 test tasks. For five agents, the Soft Q policy solves 19/20 train tasks and 18/20 test tasks. See \cref{sec:appendix_experiments} for other objectives.}
    \label{fig:real-world}
\end{figure}
Combining offline RL with real-world data should enable us to deploy our policies directly to real robots. In this experiment, we validate that our policies can succeed in the real world without any finetuning. We select the best performing hyperparameters for each loss function in \cref{sec:appendix_train_details}, and deploy the resulting policies onto either three or five robots. We evaluate each policy twice: once on train tasks, and once on test tasks (\cref{fig:real-world}, \cref{fig:demo}). We consider a task ``solved'' if the mean distance to the goal region is less than $\SI{0.25}{m}$.

In \cref{fig:real-world}, we find that the policies trained using soft and mean objectives solve most of the train tasks and \emph{generalize} to the never-before-seen tasks in the test set. Additionally, we observe zero collisions in our 40 minutes of 3-agent tests, and two minor scrapes in our 20 minutes of 5-agent tests. Upon further investigation, the scrapes appear related to issues in the motion capture system, as explained in \cref{sec:real_collisions}. Policies trained with CQL and Max Q move the agents to the edge of the boundaries and keep them there, regardless of the objective. These policies likely learn this behavior to avoid collisions between agents, but are unable to learn to navigate to the given goals. Meanwhile, policies trained using the Mean Q and Soft Q objectives reliably navigate agents to their goals.




\section{Limitations and Conclusion}
Given the complexity of fusing offline RL, LLMs, and MARL for real-world robots, we scoped this work to navigation tasks. Future work could extend our methodology towards larger and more general task-spaces. We were able to use a simple IQL learning algorithm because our navigation task has easily factorable rewards. This factorization also enabled us to utilize single-agent data. For more complex tasks with global rewards, future work could collect multi-agent datasets and train using MADDPG. This would also unlock continuous action spaces, but this would require that both the policy and Q function learn to generalize to the LLM latent space. Our random action selection might not scale to larger state spaces, however Expected SARSA Soft Q should work with any dataset collection policy.


In summary, we proposed a method that maps natural language tasks to multi-robot control outputs, by combining LLMs with offline RL. We generated multi-agent datasets from single agent data, using it to train policies offline. Then, we investigated different learning objectives, and found that safer policies work better in practice. Our experiments showed that our policies can generalize to never before seen natural language commands, and that we could deploy our policies into the real world without any finetuning.

\section{Acknowledgements}
We thank Matteo Bettini and Ryan Kortvelesy for their helpful discussions. This work was supported in part by ARL DCIST CRA W911NF-17-2-0181 and European Research Council (ERC) Project 949940 (gAIa).

\clearpage
\bibliography{bib}

\begin{thebibliography}{56}
\providecommand{\natexlab}[1]{#1}
\providecommand{\url}[1]{\texttt{#1}}
\expandafter\ifx\csname urlstyle\endcsname\relax
  \providecommand{\doi}[1]{doi: #1}\else
  \providecommand{\doi}{doi: \begingroup \urlstyle{rm}\Url}\fi

\bibitem[Vaswani et~al.(2017)Vaswani, Shazeer, Parmar, Uszkoreit, Jones, Gomez, Kaiser, and Polosukhin]{vaswani_attention_2017}
A.~Vaswani, N.~Shazeer, N.~Parmar, J.~Uszkoreit, L.~Jones, A.~N. Gomez, L.~Kaiser, and I.~Polosukhin.
\newblock Attention is {All} you {Need}.
\newblock In \emph{Advances in {Neural} {Information} {Processing} {Systems}}, volume~30. Curran Associates, Inc., 2017.
\newblock URL \url{https://proceedings.neurips.cc/paper/2017/hash/3f5ee243547dee91fbd053c1c4a845aa-Abstract.html}.

\bibitem[Brown et~al.(2020)Brown, Mann, Ryder, Subbiah, Kaplan, Dhariwal, Neelakantan, Shyam, Sastry, Askell, Agarwal, Herbert-Voss, Krueger, Henighan, Child, Ramesh, Ziegler, Wu, Winter, Hesse, Chen, Sigler, Litwin, Gray, Chess, Clark, Berner, McCandlish, Radford, Sutskever, and Amodei]{brown_language_2020}
T.~B. Brown, B.~Mann, N.~Ryder, M.~Subbiah, J.~Kaplan, P.~Dhariwal, A.~Neelakantan, P.~Shyam, G.~Sastry, A.~Askell, S.~Agarwal, A.~Herbert-Voss, G.~Krueger, T.~Henighan, R.~Child, A.~Ramesh, D.~M. Ziegler, J.~Wu, C.~Winter, C.~Hesse, M.~Chen, E.~Sigler, M.~Litwin, S.~Gray, B.~Chess, J.~Clark, C.~Berner, S.~McCandlish, A.~Radford, I.~Sutskever, and D.~Amodei.
\newblock Language models are few-shot learners.
\newblock In \emph{Advances in {Neural} {Information} {Processing} {Systems}}, volume 2020-Decem, 2020.
\newblock ISSN: 10495258 \_eprint: 2005.14165.

\bibitem[Touvron et~al.(2023)Touvron, Martin, Stone, Albert, Almahairi, Babaei, Bashlykov, Batra, Bhargava, Bhosale, Bikel, Blecher, Ferrer, Chen, Cucurull, Esiobu, Fernandes, Fu, Fu, Fuller, Gao, Goswami, Goyal, Hartshorn, Hosseini, Hou, Inan, Kardas, Kerkez, Khabsa, Kloumann, Korenev, Koura, Lachaux, Lavril, Lee, Liskovich, Lu, Mao, Martinet, Mihaylov, Mishra, Molybog, Nie, Poulton, Reizenstein, Rungta, Saladi, Schelten, Silva, Smith, Subramanian, Tan, Tang, Taylor, Williams, Kuan, Xu, Yan, Zarov, Zhang, Fan, Kambadur, Narang, Rodriguez, Stojnic, Edunov, and Scialom]{touvron_llama_2023}
H.~Touvron, L.~Martin, K.~Stone, P.~Albert, A.~Almahairi, Y.~Babaei, N.~Bashlykov, S.~Batra, P.~Bhargava, S.~Bhosale, D.~Bikel, L.~Blecher, C.~C. Ferrer, M.~Chen, G.~Cucurull, D.~Esiobu, J.~Fernandes, J.~Fu, W.~Fu, B.~Fuller, C.~Gao, V.~Goswami, N.~Goyal, A.~Hartshorn, S.~Hosseini, R.~Hou, H.~Inan, M.~Kardas, V.~Kerkez, M.~Khabsa, I.~Kloumann, A.~Korenev, P.~S. Koura, M.-A. Lachaux, T.~Lavril, J.~Lee, D.~Liskovich, Y.~Lu, Y.~Mao, X.~Martinet, T.~Mihaylov, P.~Mishra, I.~Molybog, Y.~Nie, A.~Poulton, J.~Reizenstein, R.~Rungta, K.~Saladi, A.~Schelten, R.~Silva, E.~M. Smith, R.~Subramanian, X.~E. Tan, B.~Tang, R.~Taylor, A.~Williams, J.~X. Kuan, P.~Xu, Z.~Yan, I.~Zarov, Y.~Zhang, A.~Fan, M.~Kambadur, S.~Narang, A.~Rodriguez, R.~Stojnic, S.~Edunov, and T.~Scialom.
\newblock Llama 2: {Open} {Foundation} and {Fine}-{Tuned} {Chat} {Models}, July 2023.
\newblock URL \url{http://arxiv.org/abs/2307.09288}.
\newblock arXiv:2307.09288 [cs].

\bibitem[Jiang et~al.(2023)Jiang, Sablayrolles, Mensch, Bamford, Chaplot, Casas, Bressand, Lengyel, Lample, Saulnier, Lavaud, Lachaux, Stock, Scao, Lavril, Wang, Lacroix, and Sayed]{jiang_mistral_2023}
A.~Q. Jiang, A.~Sablayrolles, A.~Mensch, C.~Bamford, D.~S. Chaplot, D.~d.~l. Casas, F.~Bressand, G.~Lengyel, G.~Lample, L.~Saulnier, L.~R. Lavaud, M.-A. Lachaux, P.~Stock, T.~L. Scao, T.~Lavril, T.~Wang, T.~Lacroix, and W.~E. Sayed.
\newblock Mistral {7B}, Oct. 2023.
\newblock URL \url{http://arxiv.org/abs/2310.06825}.
\newblock arXiv:2310.06825 [cs].

\bibitem[OpenAI et~al.(2024)OpenAI, Achiam, Adler, Agarwal, Ahmad, Akkaya, Aleman, Almeida, Altenschmidt, Altman, Anadkat, Avila, Babuschkin, Balaji, Balcom, Baltescu, Bao, Bavarian, Belgum, Bello, Berdine, Bernadett-Shapiro, Berner, Bogdonoff, Boiko, Boyd, Brakman, Brockman, Brooks, Brundage, Button, Cai, Campbell, Cann, Carey, Carlson, Carmichael, Chan, Chang, Chantzis, Chen, Chen, Chen, Chen, Chen, Chess, Cho, Chu, Chung, Cummings, Currier, Dai, Decareaux, Degry, Deutsch, Deville, Dhar, Dohan, Dowling, Dunning, Ecoffet, Eleti, Eloundou, Farhi, Fedus, Felix, Fishman, Forte, Fulford, Gao, Georges, Gibson, Goel, Gogineni, Goh, Gontijo-Lopes, Gordon, Grafstein, Gray, Greene, Gross, Gu, Guo, Hallacy, Han, Harris, He, Heaton, Heidecke, Hesse, Hickey, Hickey, Hoeschele, Houghton, Hsu, Hu, Hu, Huizinga, Jain, Jain, Jang, Jiang, Jiang, Jin, Jin, Jomoto, Jonn, Jun, Kaftan, Kaiser, Kamali, Kanitscheider, Keskar, Khan, Kilpatrick, Kim, Kim, Kim, Kirchner, Kiros, Knight, Kokotajlo, Kondraciuk, Kondrich, Konstantinidis,
  Kosic, Krueger, Kuo, Lampe, Lan, Lee, Leike, Leung, Levy, Li, Lim, Lin, Lin, Litwin, Lopez, Lowe, Lue, Makanju, Malfacini, Manning, Markov, Markovski, Martin, Mayer, Mayne, McGrew, McKinney, McLeavey, McMillan, McNeil, Medina, Mehta, Menick, Metz, Mishchenko, Mishkin, Monaco, Morikawa, Mossing, Mu, Murati, Murk, Mély, Nair, Nakano, Nayak, Neelakantan, Ngo, Noh, Ouyang, O'Keefe, Pachocki, Paino, Palermo, Pantuliano, Parascandolo, Parish, Parparita, Passos, Pavlov, Peng, Perelman, Peres, Petrov, Pinto, Michael, Pokorny, Pokrass, Pong, Powell, Power, Power, Proehl, Puri, Radford, Rae, Ramesh, Raymond, Real, Rimbach, Ross, Rotsted, Roussez, Ryder, Saltarelli, Sanders, Santurkar, Sastry, Schmidt, Schnurr, Schulman, Selsam, Sheppard, Sherbakov, Shieh, Shoker, Shyam, Sidor, Sigler, Simens, Sitkin, Slama, Sohl, Sokolowsky, Song, Staudacher, Such, Summers, Sutskever, Tang, Tezak, Thompson, Tillet, Tootoonchian, Tseng, Tuggle, Turley, Tworek, Uribe, Vallone, Vijayvergiya, Voss, Wainwright, Wang, Wang, Wang, Ward,
  Wei, Weinmann, Welihinda, Welinder, Weng, Weng, Wiethoff, Willner, Winter, Wolrich, Wong, Workman, Wu, Wu, Wu, Xiao, Xu, Yoo, Yu, Yuan, Zaremba, Zellers, Zhang, Zhang, Zhao, Zheng, Zhuang, Zhuk, and Zoph]{openai_gpt-4_2024}
OpenAI, J.~Achiam, S.~Adler, S.~Agarwal, L.~Ahmad, I.~Akkaya, F.~L. Aleman, D.~Almeida, J.~Altenschmidt, S.~Altman, S.~Anadkat, R.~Avila, I.~Babuschkin, S.~Balaji, V.~Balcom, P.~Baltescu, H.~Bao, M.~Bavarian, J.~Belgum, I.~Bello, J.~Berdine, G.~Bernadett-Shapiro, C.~Berner, L.~Bogdonoff, O.~Boiko, M.~Boyd, A.-L. Brakman, G.~Brockman, T.~Brooks, M.~Brundage, K.~Button, T.~Cai, R.~Campbell, A.~Cann, B.~Carey, C.~Carlson, R.~Carmichael, B.~Chan, C.~Chang, F.~Chantzis, D.~Chen, S.~Chen, R.~Chen, J.~Chen, M.~Chen, B.~Chess, C.~Cho, C.~Chu, H.~W. Chung, D.~Cummings, J.~Currier, Y.~Dai, C.~Decareaux, T.~Degry, N.~Deutsch, D.~Deville, A.~Dhar, D.~Dohan, S.~Dowling, S.~Dunning, A.~Ecoffet, A.~Eleti, T.~Eloundou, D.~Farhi, L.~Fedus, N.~Felix, S.~P. Fishman, J.~Forte, I.~Fulford, L.~Gao, E.~Georges, C.~Gibson, V.~Goel, T.~Gogineni, G.~Goh, R.~Gontijo-Lopes, J.~Gordon, M.~Grafstein, S.~Gray, R.~Greene, J.~Gross, S.~S. Gu, Y.~Guo, C.~Hallacy, J.~Han, J.~Harris, Y.~He, M.~Heaton, J.~Heidecke, C.~Hesse, A.~Hickey,
  W.~Hickey, P.~Hoeschele, B.~Houghton, K.~Hsu, S.~Hu, X.~Hu, J.~Huizinga, S.~Jain, S.~Jain, J.~Jang, A.~Jiang, R.~Jiang, H.~Jin, D.~Jin, S.~Jomoto, B.~Jonn, H.~Jun, T.~Kaftan, L.~Kaiser, A.~Kamali, I.~Kanitscheider, N.~S. Keskar, T.~Khan, L.~Kilpatrick, J.~W. Kim, C.~Kim, Y.~Kim, J.~H. Kirchner, J.~Kiros, M.~Knight, D.~Kokotajlo, L.~Kondraciuk, A.~Kondrich, A.~Konstantinidis, K.~Kosic, G.~Krueger, V.~Kuo, M.~Lampe, I.~Lan, T.~Lee, J.~Leike, J.~Leung, D.~Levy, C.~M. Li, R.~Lim, M.~Lin, S.~Lin, M.~Litwin, T.~Lopez, R.~Lowe, P.~Lue, A.~Makanju, K.~Malfacini, S.~Manning, T.~Markov, Y.~Markovski, B.~Martin, K.~Mayer, A.~Mayne, B.~McGrew, S.~M. McKinney, C.~McLeavey, P.~McMillan, J.~McNeil, D.~Medina, A.~Mehta, J.~Menick, L.~Metz, A.~Mishchenko, P.~Mishkin, V.~Monaco, E.~Morikawa, D.~Mossing, T.~Mu, M.~Murati, O.~Murk, D.~Mély, A.~Nair, R.~Nakano, R.~Nayak, A.~Neelakantan, R.~Ngo, H.~Noh, L.~Ouyang, C.~O'Keefe, J.~Pachocki, A.~Paino, J.~Palermo, A.~Pantuliano, G.~Parascandolo, J.~Parish, E.~Parparita, A.~Passos,
  M.~Pavlov, A.~Peng, A.~Perelman, F.~d. A.~B. Peres, M.~Petrov, H.~P. d.~O. Pinto, Michael, Pokorny, M.~Pokrass, V.~H. Pong, T.~Powell, A.~Power, B.~Power, E.~Proehl, R.~Puri, A.~Radford, J.~Rae, A.~Ramesh, C.~Raymond, F.~Real, K.~Rimbach, C.~Ross, B.~Rotsted, H.~Roussez, N.~Ryder, M.~Saltarelli, T.~Sanders, S.~Santurkar, G.~Sastry, H.~Schmidt, D.~Schnurr, J.~Schulman, D.~Selsam, K.~Sheppard, T.~Sherbakov, J.~Shieh, S.~Shoker, P.~Shyam, S.~Sidor, E.~Sigler, M.~Simens, J.~Sitkin, K.~Slama, I.~Sohl, B.~Sokolowsky, Y.~Song, N.~Staudacher, F.~P. Such, N.~Summers, I.~Sutskever, J.~Tang, N.~Tezak, M.~B. Thompson, P.~Tillet, A.~Tootoonchian, E.~Tseng, P.~Tuggle, N.~Turley, J.~Tworek, J.~F.~C. Uribe, A.~Vallone, A.~Vijayvergiya, C.~Voss, C.~Wainwright, J.~J. Wang, A.~Wang, B.~Wang, J.~Ward, J.~Wei, C.~J. Weinmann, A.~Welihinda, P.~Welinder, J.~Weng, L.~Weng, M.~Wiethoff, D.~Willner, C.~Winter, S.~Wolrich, H.~Wong, L.~Workman, S.~Wu, J.~Wu, M.~Wu, K.~Xiao, T.~Xu, S.~Yoo, K.~Yu, Q.~Yuan, W.~Zaremba, R.~Zellers,
  C.~Zhang, M.~Zhang, S.~Zhao, T.~Zheng, J.~Zhuang, W.~Zhuk, and B.~Zoph.
\newblock {GPT}-4 {Technical} {Report}, Mar. 2024.
\newblock URL \url{http://arxiv.org/abs/2303.08774}.
\newblock arXiv:2303.08774 [cs].

\bibitem[Driess et~al.(2023)Driess, Xia, Sajjadi, Lynch, Chowdhery, Ichter, Wahid, Tompson, Vuong, Yu, Huang, Chebotar, Sermanet, Duckworth, Levine, Vanhoucke, Hausman, Toussaint, Greff, Zeng, Mordatch, and Florence]{driess_palm-e_2023}
D.~Driess, F.~Xia, M.~S.~M. Sajjadi, C.~Lynch, A.~Chowdhery, B.~Ichter, A.~Wahid, J.~Tompson, Q.~Vuong, T.~Yu, W.~Huang, Y.~Chebotar, P.~Sermanet, D.~Duckworth, S.~Levine, V.~Vanhoucke, K.~Hausman, M.~Toussaint, K.~Greff, A.~Zeng, I.~Mordatch, and P.~Florence.
\newblock {PaLM}-{E}: {An} {Embodied} {Multimodal} {Language} {Model}, Mar. 2023.
\newblock URL \url{http://arxiv.org/abs/2303.03378}.
\newblock arXiv:2303.03378 [cs].

\bibitem[Dorbala et~al.(2024)Dorbala, Mullen, and Manocha]{dorbala_can_2024}
V.~S. Dorbala, J.~F. Mullen, and D.~Manocha.
\newblock Can an {Embodied} {Agent} {Find} {Your} “{Cat}-shaped {Mug}”? {LLM}-{Based} {Zero}-{Shot} {Object} {Navigation}.
\newblock \emph{IEEE Robotics and Automation Letters}, 9\penalty0 (5):\penalty0 4083--4090, May 2024.
\newblock ISSN 2377-3766.
\newblock \doi{10.1109/LRA.2023.3346800}.
\newblock URL \url{https://ieeexplore.ieee.org/document/10373065}.
\newblock Conference Name: IEEE Robotics and Automation Letters.

\bibitem[Brohan et~al.(2023)Brohan, Brown, Carbajal, Chebotar, Chen, Choromanski, Ding, Driess, Dubey, Finn, Florence, Fu, Arenas, Gopalakrishnan, Han, Hausman, Herzog, Hsu, Ichter, Irpan, Joshi, Julian, Kalashnikov, Kuang, Leal, Lee, Lee, Levine, Lu, Michalewski, Mordatch, Pertsch, Rao, Reymann, Ryoo, Salazar, Sanketi, Sermanet, Singh, Singh, Soricut, Tran, Vanhoucke, Vuong, Wahid, Welker, Wohlhart, Wu, Xia, Xiao, Xu, Xu, Yu, and Zitkovich]{brohan_rt-2_2023}
A.~Brohan, N.~Brown, J.~Carbajal, Y.~Chebotar, X.~Chen, K.~Choromanski, T.~Ding, D.~Driess, A.~Dubey, C.~Finn, P.~Florence, C.~Fu, M.~G. Arenas, K.~Gopalakrishnan, K.~Han, K.~Hausman, A.~Herzog, J.~Hsu, B.~Ichter, A.~Irpan, N.~Joshi, R.~Julian, D.~Kalashnikov, Y.~Kuang, I.~Leal, L.~Lee, T.-W.~E. Lee, S.~Levine, Y.~Lu, H.~Michalewski, I.~Mordatch, K.~Pertsch, K.~Rao, K.~Reymann, M.~Ryoo, G.~Salazar, P.~Sanketi, P.~Sermanet, J.~Singh, A.~Singh, R.~Soricut, H.~Tran, V.~Vanhoucke, Q.~Vuong, A.~Wahid, S.~Welker, P.~Wohlhart, J.~Wu, F.~Xia, T.~Xiao, P.~Xu, S.~Xu, T.~Yu, and B.~Zitkovich.
\newblock {RT}-2: {Vision}-{Language}-{Action} {Models} {Transfer} {Web} {Knowledge} to {Robotic} {Control}, July 2023.
\newblock URL \url{http://arxiv.org/abs/2307.15818}.
\newblock arXiv:2307.15818 [cs].

\bibitem[Shah et~al.(2023)Shah, Osiński, Ichter, and Levine]{shah_lm-nav_2023}
D.~Shah, B.~Osiński, B.~Ichter, and S.~Levine.
\newblock {LM}-{Nav}: {Robotic} {Navigation} with {Large} {Pre}-{Trained} {Models} of {Language}, {Vision}, and {Action}.
\newblock In \emph{Proceedings of {The} 6th {Conference} on {Robot} {Learning}}, pages 492--504. PMLR, Mar. 2023.
\newblock URL \url{https://proceedings.mlr.press/v205/shah23b.html}.
\newblock ISSN: 2640-3498.

\bibitem[Ni and Yang(2011)]{ni_bioinspired_2011}
J.~Ni and S.~X. Yang.
\newblock Bioinspired {Neural} {Network} for {Real}-{Time} {Cooperative} {Hunting} by {Multirobots} in {Unknown} {Environments}.
\newblock \emph{IEEE Transactions on Neural Networks}, 22\penalty0 (12):\penalty0 2062--2077, Dec. 2011.
\newblock ISSN 1941-0093.
\newblock \doi{10.1109/TNN.2011.2169808}.
\newblock URL \url{https://ieeexplore.ieee.org/abstract/document/6060919?casa_token=0BzAmcnZeKAAAAAA:W8_-gpjhR5KRIqNbwfk4S0kwK-v002YCxxwUNCBDlRpT17pNxDc9IF-NgQyTzImKV1RjFdlh}.
\newblock Conference Name: IEEE Transactions on Neural Networks.

\bibitem[Chen et~al.(2017)Chen, Liu, Everett, and How]{chen_decentralized_2017}
Y.~F. Chen, M.~Liu, M.~Everett, and J.~P. How.
\newblock Decentralized non-communicating multiagent collision avoidance with deep reinforcement learning.
\newblock In \emph{2017 {IEEE} {International} {Conference} on {Robotics} and {Automation} ({ICRA})}, pages 285--292, May 2017.
\newblock \doi{10.1109/ICRA.2017.7989037}.
\newblock URL \url{https://ieeexplore.ieee.org/abstract/document/7989037?casa_token=RStkd0_OxqIAAAAA:aiuKOZ45U5RxHqXdJBPcizCV4U9acJIlJtw9uajpTqCnnPedB2j408iN4rnPA-xJv6GTr2_Z}.

\bibitem[Blumenkamp et~al.(2022)Blumenkamp, Morad, Gielis, Li, and Prorok]{blumenkamp_framework_2022}
J.~Blumenkamp, S.~Morad, J.~Gielis, Q.~Li, and A.~Prorok.
\newblock A {Framework} for {Real}-{World} {Multi}-{Robot} {Systems} {Running} {Decentralized} {GNN}-{Based} {Policies}.
\newblock In \emph{2022 {International} {Conference} on {Robotics} and {Automation} ({ICRA})}, pages 8772--8778, 2022.
\newblock \doi{10.1109/ICRA46639.2022.9811744}.

\bibitem[Lynch et~al.(2023)Lynch, Wahid, Tompson, Ding, Betker, Baruch, Armstrong, and Florence]{lynch_interactive_2023}
C.~Lynch, A.~Wahid, J.~Tompson, T.~Ding, J.~Betker, R.~Baruch, T.~Armstrong, and P.~Florence.
\newblock Interactive {Language}: {Talking} to {Robots} in {Real} {Time}.
\newblock \emph{IEEE Robotics and Automation Letters}, pages 1--8, 2023.
\newblock ISSN 2377-3766.
\newblock \doi{10.1109/LRA.2023.3295255}.
\newblock URL \url{https://ieeexplore.ieee.org/document/10182264}.
\newblock Conference Name: IEEE Robotics and Automation Letters.

\bibitem[Ichter et~al.(2023)Ichter, Brohan, Chebotar, Finn, Hausman, Herzog, Ho, Ibarz, Irpan, Jang, Julian, Kalashnikov, Levine, Lu, Parada, Rao, Sermanet, Toshev, Vanhoucke, Xia, Xiao, Xu, Yan, Brown, Ahn, Cortes, Sievers, Tan, Xu, Reyes, Rettinghouse, Quiambao, Pastor, Luu, Lee, Kuang, Jesmonth, Joshi, Jeffrey, Ruano, Hsu, Gopalakrishnan, David, Zeng, and Fu]{ichter_as_2023}
B.~Ichter, A.~Brohan, Y.~Chebotar, C.~Finn, K.~Hausman, A.~Herzog, D.~Ho, J.~Ibarz, A.~Irpan, E.~Jang, R.~Julian, D.~Kalashnikov, S.~Levine, Y.~Lu, C.~Parada, K.~Rao, P.~Sermanet, A.~T. Toshev, V.~Vanhoucke, F.~Xia, T.~Xiao, P.~Xu, M.~Yan, N.~Brown, M.~Ahn, O.~Cortes, N.~Sievers, C.~Tan, S.~Xu, D.~Reyes, J.~Rettinghouse, J.~Quiambao, P.~Pastor, L.~Luu, K.-H. Lee, Y.~Kuang, S.~Jesmonth, N.~J. Joshi, K.~Jeffrey, R.~J. Ruano, J.~Hsu, K.~Gopalakrishnan, B.~David, A.~Zeng, and C.~K. Fu.
\newblock Do {As} {I} {Can}, {Not} {As} {I} {Say}: {Grounding} {Language} in {Robotic} {Affordances}.
\newblock In \emph{Proceedings of {The} 6th {Conference} on {Robot} {Learning}}, pages 287--318. PMLR, Mar. 2023.
\newblock URL \url{https://proceedings.mlr.press/v205/ichter23a.html}.
\newblock ISSN: 2640-3498.

\bibitem[{DFRobot}(2023)]{dfrobot_nvidia_2023}
{DFRobot}.
\newblock {NVIDIA} {Jetson} {AGX} {Orin} {Large} {Language} {Model} {LLaMA2}-7b and {LLaMA2}-13b {Performance} {Test} {Report}, 2023.
\newblock URL \url{https://www.dfrobot.com}.

\bibitem[Kannan et~al.(2024)Kannan, Venkatesh, and Min]{kannan_smart-llm_2024}
S.~S. Kannan, V.~L.~N. Venkatesh, and B.-C. Min.
\newblock {SMART}-{LLM}: {Smart} {Multi}-{Agent} {Robot} {Task} {Planning} using {Large} {Language} {Models}, Mar. 2024.
\newblock URL \url{http://arxiv.org/abs/2309.10062}.
\newblock arXiv:2309.10062 [cs].

\bibitem[Mandi et~al.(2023)Mandi, Jain, and Song]{mandi_roco_2023}
Z.~Mandi, S.~Jain, and S.~Song.
\newblock {RoCo}: {Dialectic} {Multi}-{Robot} {Collaboration} with {Large} {Language} {Models}, July 2023.
\newblock URL \url{http://arxiv.org/abs/2307.04738}.
\newblock arXiv:2307.04738 [cs].

\bibitem[Chan et~al.(2023)Chan, Chen, Su, Yu, Xue, Zhang, Fu, and Liu]{chan_chateval_2023}
C.-M. Chan, W.~Chen, Y.~Su, J.~Yu, W.~Xue, S.~Zhang, J.~Fu, and Z.~Liu.
\newblock {ChatEval}: {Towards} {Better} {LLM}-based {Evaluators} through {Multi}-{Agent} {Debate}.
\newblock Oct. 2023.
\newblock URL \url{https://openreview.net/forum?id=FQepisCUWu}.

\bibitem[Zhang et~al.(2023)Zhang, Du, Shan, Zhou, Du, Tenenbaum, Shu, and Gan]{zhang_building_2023}
H.~Zhang, W.~Du, J.~Shan, Q.~Zhou, Y.~Du, J.~B. Tenenbaum, T.~Shu, and C.~Gan.
\newblock Building {Cooperative} {Embodied} {Agents} {Modularly} with {Large} {Language} {Models}.
\newblock Oct. 2023.
\newblock URL \url{https://openreview.net/forum?id=EnXJfQqy0K}.

\bibitem[Garg et~al.(2024)Garg, Arkin, Zhang, Roy, and Fan]{garg_large_2024}
K.~Garg, J.~Arkin, S.~Zhang, N.~Roy, and C.~Fan.
\newblock Large {Language} {Models} to the {Rescue}: {Deadlock} {Resolution} in {Multi}-{Robot} {Systems}, Apr. 2024.
\newblock URL \url{http://arxiv.org/abs/2404.06413}.
\newblock arXiv:2404.06413 [cs, math].

\bibitem[Reimers and Gurevych(2019)]{reimers_sentence-bert_2019}
N.~Reimers and I.~Gurevych.
\newblock Sentence-{BERT}: {Sentence} {Embeddings} using {Siamese} {BERT}-{Networks}.
\newblock In \emph{Proceedings of the 2019 {Conference} on {Empirical} {Methods} in {Natural} {Language} {Processing}}. Association for Computational Linguistics, Nov. 2019.
\newblock URL \url{https://arxiv.org/abs/1908.10084}.

\bibitem[Li et~al.(2023)Li, Zhang, Zhang, Long, Xie, and Zhang]{li_towards_2023}
Z.~Li, X.~Zhang, Y.~Zhang, D.~Long, P.~Xie, and M.~Zhang.
\newblock Towards {General} {Text} {Embeddings} with {Multi}-stage {Contrastive} {Learning}, Aug. 2023.
\newblock URL \url{http://arxiv.org/abs/2308.03281}.
\newblock arXiv:2308.03281 [cs].

\bibitem[Lee et~al.(2024)Lee, Roy, Xu, Raiman, Shoeybi, Catanzaro, and Ping]{lee_nv-embed_2024}
C.~Lee, R.~Roy, M.~Xu, J.~Raiman, M.~Shoeybi, B.~Catanzaro, and W.~Ping.
\newblock {NV}-{Embed}: {Improved} {Techniques} for {Training} {LLMs} as {Generalist} {Embedding} {Models}, May 2024.
\newblock URL \url{http://arxiv.org/abs/2405.17428}.
\newblock arXiv:2405.17428 [cs].

\bibitem[Sutton and Barto(2018)]{sutton_reinforcement_2018}
R.~S. Sutton and A.~G. Barto.
\newblock \emph{Reinforcement learning: {An} introduction}.
\newblock MIT press, 2018.

\bibitem[Yang and Wang(2021)]{yang_overview_2021}
Y.~Yang and J.~Wang.
\newblock An {Overview} of {Multi}-{Agent} {Reinforcement} {Learning} from {Game} {Theoretical} {Perspective}, Mar. 2021.
\newblock URL \url{http://arxiv.org/abs/2011.00583}.
\newblock arXiv:2011.00583 [cs].

\bibitem[Lowe et~al.(2017)Lowe, Wu, Tamar, Harb, Abbeel, and Mordatch]{lowe_multi-agent_2017}
R.~Lowe, Y.~Wu, A.~Tamar, J.~Harb, P.~Abbeel, and I.~Mordatch.
\newblock Multi-agent actor-critic for mixed cooperative-competitive environments.
\newblock In \emph{Advances in {Neural} {Information} {Processing} {Systems}}, volume 2017-Decem, 2017.
\newblock ISSN: 10495258.

\bibitem[Rashid et~al.(2020)Rashid, Samvelyan, Witt, Farquhar, Foerster, and Whiteson]{rashid_monotonic_2020}
T.~Rashid, M.~Samvelyan, C.~S.~d. Witt, G.~Farquhar, J.~Foerster, and S.~Whiteson.
\newblock Monotonic {Value} {Function} {Factorisation} for {Deep} {Multi}-{Agent} {Reinforcement} {Learning}.
\newblock \emph{Journal of Machine Learning Research}, 21\penalty0 (178):\penalty0 1--51, 2020.
\newblock ISSN 1533-7928.
\newblock URL \url{http://jmlr.org/papers/v21/20-081.html}.

\bibitem[Foerster et~al.(2018)Foerster, Farquhar, Afouras, Nardelli, and Whiteson]{foerster_counterfactual_2018}
J.~Foerster, G.~Farquhar, T.~Afouras, N.~Nardelli, and S.~Whiteson.
\newblock Counterfactual {Multi}-{Agent} {Policy} {Gradients}.
\newblock \emph{Proceedings of the AAAI Conference on Artificial Intelligence}, 32\penalty0 (1), Apr. 2018.
\newblock ISSN 2374-3468.
\newblock \doi{10.1609/aaai.v32i1.11794}.
\newblock URL \url{https://ojs.aaai.org/index.php/AAAI/article/view/11794}.
\newblock Number: 1.

\bibitem[de~Witt et~al.(2020)de~Witt, Gupta, Makoviichuk, Makoviychuk, Torr, Sun, and Whiteson]{de_witt_is_2020}
C.~S. de~Witt, T.~Gupta, D.~Makoviichuk, V.~Makoviychuk, P.~H.~S. Torr, M.~Sun, and S.~Whiteson.
\newblock Is {Independent} {Learning} {All} {You} {Need} in the {StarCraft} {Multi}-{Agent} {Challenge}?, Nov. 2020.
\newblock URL \url{http://arxiv.org/abs/2011.09533}.
\newblock arXiv:2011.09533 [cs].

\bibitem[Tan(1993)]{tan_multi-agent_1993}
M.~Tan.
\newblock Multi-agent reinforcement learning: {Independent} vs. cooperative agents.
\newblock In \emph{Proceedings of the tenth international conference on machine learning}, pages 330--337, 1993.

\bibitem[Pan et~al.(2021)Pan, Rashid, Peng, Huang, and Whiteson]{pan_regularized_2021}
L.~Pan, T.~Rashid, B.~Peng, L.~Huang, and S.~Whiteson.
\newblock Regularized {Softmax} {Deep} {Multi}-{Agent} {Q}-{Learning}.
\newblock In \emph{Advances in {Neural} {Information} {Processing} {Systems}}, volume~34, pages 1365--1377. Curran Associates, Inc., 2021.
\newblock URL \url{https://proceedings.neurips.cc/paper/2021/hash/0a113ef6b61820daa5611c870ed8d5ee-Abstract.html}.

\bibitem[Kumar et~al.(2020)Kumar, Zhou, Tucker, and Levine]{kumar_conservative_2020}
A.~Kumar, A.~Zhou, G.~Tucker, and S.~Levine.
\newblock Conservative {Q}-{Learning} for {Offline} {Reinforcement} {Learning}.
\newblock In \emph{Advances in {Neural} {Information} {Processing} {Systems}}, volume~33, pages 1179--1191. Curran Associates, Inc., 2020.
\newblock URL \url{https://proceedings.neurips.cc/paper/2020/hash/0d2b2061826a5df3221116a5085a6052-Abstract.html}.

\bibitem[Kumar et~al.(2019)Kumar, Fu, Soh, Tucker, and Levine]{kumar_stabilizing_2019}
A.~Kumar, J.~Fu, M.~Soh, G.~Tucker, and S.~Levine.
\newblock Stabilizing {Off}-{Policy} {Q}-{Learning} via {Bootstrapping} {Error} {Reduction}.
\newblock In \emph{Advances in {Neural} {Information} {Processing} {Systems}}, volume~32. Curran Associates, Inc., 2019.
\newblock URL \url{https://proceedings.neurips.cc/paper_files/paper/2019/hash/c2073ffa77b5357a498057413bb09d3a-Abstract.html}.

\bibitem[Shao et~al.(2023)Shao, Qu, Chen, Zhang, and Ji]{shao_counterfactual_2023}
J.~Shao, Y.~Qu, C.~Chen, H.~Zhang, and X.~Ji.
\newblock Counterfactual {Conservative} {Q} {Learning} for {Offline} {Multi}-agent {Reinforcement} {Learning}.
\newblock \emph{Advances in Neural Information Processing Systems}, 36:\penalty0 77290--77312, Dec. 2023.
\newblock URL \url{https://proceedings.neurips.cc/paper_files/paper/2023/hash/f3f2ff9579ba6deeb89caa2fe1f0b99c-Abstract-Conference.html}.

\bibitem[Fujimoto and Gu(2021)]{fujimoto_minimalist_2021}
S.~Fujimoto and S.~S. Gu.
\newblock A {Minimalist} {Approach} to {Offline} {Reinforcement} {Learning}.
\newblock In \emph{Advances in {Neural} {Information} {Processing} {Systems}}, volume~34, pages 20132--20145. Curran Associates, Inc., 2021.
\newblock URL \url{https://proceedings.neurips.cc/paper_files/paper/2021/hash/a8166da05c5a094f7dc03724b41886e5-Abstract.html}.

\bibitem[Ball et~al.(2023)Ball, Smith, Kostrikov, and Levine]{ball_efficient_2023}
P.~J. Ball, L.~Smith, I.~Kostrikov, and S.~Levine.
\newblock Efficient {Online} {Reinforcement} {Learning} with {Offline} {Data}.
\newblock In \emph{Proceedings of the 40th {International} {Conference} on {Machine} {Learning}}, pages 1577--1594. PMLR, July 2023.
\newblock URL \url{https://proceedings.mlr.press/v202/ball23a.html}.
\newblock ISSN: 2640-3498.

\bibitem[Yang et~al.(2021)Yang, Ma, Li, Zheng, Zhang, Huang, Yang, and Zhao]{yang_believe_2021}
Y.~Yang, X.~Ma, C.~Li, Z.~Zheng, Q.~Zhang, G.~Huang, J.~Yang, and Q.~Zhao.
\newblock Believe {What} {You} {See}: {Implicit} {Constraint} {Approach} for {Offline} {Multi}-{Agent} {Reinforcement} {Learning}.
\newblock In \emph{Advances in {Neural} {Information} {Processing} {Systems}}, volume~34, pages 10299--10312. Curran Associates, Inc., 2021.
\newblock URL \url{https://proceedings.neurips.cc/paper/2021/hash/550a141f12de6341fba65b0ad0433500-Abstract.html}.

\bibitem[Liu et~al.(2022)Liu, Zhu, and Zhang]{liu_goal-conditioned_2022}
M.~Liu, M.~Zhu, and W.~Zhang.
\newblock Goal-{Conditioned} {Reinforcement} {Learning}: {Problems} and {Solutions}.
\newblock In \emph{Proceedings of the {Thirty}-{First} {International} {Joint} {Conference} on {Artificial} {Intelligence}}, pages 5502--5511, Vienna, Austria, July 2022. International Joint Conferences on Artificial Intelligence Organization.
\newblock ISBN 978-1-956792-00-3.
\newblock \doi{10.24963/ijcai.2022/770}.
\newblock URL \url{https://www.ijcai.org/proceedings/2022/770}.

\bibitem[Teh et~al.(2017)Teh, Bapst, Czarnecki, Quan, Kirkpatrick, Hadsell, Heess, and Pascanu]{teh_distral_2017}
Y.~Teh, V.~Bapst, W.~M. Czarnecki, J.~Quan, J.~Kirkpatrick, R.~Hadsell, N.~Heess, and R.~Pascanu.
\newblock Distral: {Robust} multitask reinforcement learning.
\newblock In \emph{Advances in {Neural} {Information} {Processing} {Systems}}, volume~30. Curran Associates, Inc., 2017.
\newblock URL \url{https://proceedings.neurips.cc/paper_files/paper/2017/hash/0abdc563a06105aee3c6136871c9f4d1-Abstract.html}.

\bibitem[Schaul et~al.(2015)Schaul, Horgan, Gregor, and Silver]{schaul_universal_2015}
T.~Schaul, D.~Horgan, K.~Gregor, and D.~Silver.
\newblock Universal {Value} {Function} {Approximators}.
\newblock In \emph{Proceedings of the 32nd {International} {Conference} on {Machine} {Learning}}, pages 1312--1320. PMLR, June 2015.
\newblock URL \url{https://proceedings.mlr.press/v37/schaul15.html}.
\newblock ISSN: 1938-7228.

\bibitem[Blumenkamp et~al.(2024)Blumenkamp, Shankar, Bettini, Bird, and Prorok]{blumenkamp_cambridge_2024}
J.~Blumenkamp, A.~Shankar, M.~Bettini, J.~Bird, and A.~Prorok.
\newblock The {Cambridge} {RoboMaster}: {An} {Agile} {Multi}-{Robot} {Research} {Platform}, May 2024.
\newblock URL \url{http://arxiv.org/abs/2405.02198}.
\newblock arXiv:2405.02198 [cs, eess].

\bibitem[Bettini et~al.(2022)Bettini, Kortvelesy, Blumenkamp, and Prorok]{bettini_vmas_2022}
M.~Bettini, R.~Kortvelesy, J.~Blumenkamp, and A.~Prorok.
\newblock {VMAS}: {A} {Vectorized} {Multi}-{Agent} {Simulator} for {Collective} {Robot} {Learning}.
\newblock \emph{The 16th International Symposium on Distributed Autonomous Robotic Systems}, 2022.
\newblock Publisher: Springer.

\bibitem[van Seijen et~al.(2009)van Seijen, van Hasselt, Whiteson, and Wiering]{van_seijen_theoretical_2009}
H.~van Seijen, H.~van Hasselt, S.~Whiteson, and M.~Wiering.
\newblock A theoretical and empirical analysis of {Expected} {Sarsa}.
\newblock In \emph{2009 {IEEE} {Symposium} on {Adaptive} {Dynamic} {Programming} and {Reinforcement} {Learning}}, pages 177--184, Mar. 2009.
\newblock \doi{10.1109/ADPRL.2009.4927542}.
\newblock URL \url{https://ieeexplore.ieee.org/abstract/document/4927542}.
\newblock ISSN: 2325-1867.

\bibitem[Emmons et~al.(2021)Emmons, Eysenbach, Kostrikov, and Levine]{emmons_rvs_2021}
S.~Emmons, B.~Eysenbach, I.~Kostrikov, and S.~Levine.
\newblock {RvS}: {What} is {Essential} for {Offline} {RL} via {Supervised} {Learning}?
\newblock Oct. 2021.
\newblock URL \url{https://openreview.net/forum?id=S874XAIpkR-}.

\bibitem[Ba et~al.(2016)Ba, Kiros, and Hinton]{ba_layer_2016}
J.~L. Ba, J.~R. Kiros, and G.~E. Hinton.
\newblock Layer {Normalization}.
\newblock July 2016.
\newblock URL \url{https://arxiv.org/abs/1607.06450v1 http://arxiv.org/abs/1607.06450}.
\newblock \_eprint: 1607.06450.

\bibitem[Xu et~al.(2015)Xu, Wang, Chen, and Li]{xu_empirical_2015}
B.~Xu, N.~Wang, T.~Chen, and M.~Li.
\newblock Empirical evaluation of rectified activations in convolutional network.
\newblock \emph{arXiv preprint arXiv:1505.00853}, 2015.

\bibitem[Wang et~al.(2016)Wang, Schaul, Hessel, Hasselt, Lanctot, and Freitas]{wang_dueling_2016}
Z.~Wang, T.~Schaul, M.~Hessel, H.~Hasselt, M.~Lanctot, and N.~Freitas.
\newblock Dueling {Network} {Architectures} for {Deep} {Reinforcement} {Learning}.
\newblock In \emph{Proceedings of {The} 33rd {International} {Conference} on {Machine} {Learning}}, pages 1995--2003. PMLR, June 2016.
\newblock URL \url{https://proceedings.mlr.press/v48/wangf16.html}.
\newblock ISSN: 1938-7228.

\bibitem[Fujimoto et~al.(2018)Fujimoto, Hoof, and Meger]{fujimoto_addressing_2018}
S.~Fujimoto, H.~Hoof, and D.~Meger.
\newblock Addressing {Function} {Approximation} {Error} in {Actor}-{Critic} {Methods}.
\newblock In \emph{Proceedings of the 35th {International} {Conference} on {Machine} {Learning}}, pages 1587--1596. PMLR, July 2018.
\newblock URL \url{https://proceedings.mlr.press/v80/fujimoto18a.html}.
\newblock ISSN: 2640-3498.

\bibitem[Li and Li(2023)]{li_angle-optimized_2023}
X.~Li and J.~Li.
\newblock {AnglE}-optimized {Text} {Embeddings}.
\newblock \emph{arXiv preprint arXiv:2309.12871}, 2023.

\bibitem[Feng et~al.(2022)Feng, Yang, Cer, Arivazhagan, and Wang]{feng_language-agnostic_2022}
F.~Feng, Y.~Yang, D.~Cer, N.~Arivazhagan, and W.~Wang.
\newblock Language-agnostic {BERT} {Sentence} {Embedding}.
\newblock In S.~Muresan, P.~Nakov, and A.~Villavicencio, editors, \emph{Proceedings of the 60th {Annual} {Meeting} of the {Association} for {Computational} {Linguistics} ({Volume} 1: {Long} {Papers})}, pages 878--891, Dublin, Ireland, May 2022. Association for Computational Linguistics.
\newblock \doi{10.18653/v1/2022.acl-long.62}.
\newblock URL \url{https://aclanthology.org/2022.acl-long.62}.

\bibitem[Muennighoff et~al.(2023)Muennighoff, Tazi, Magne, and Reimers]{muennighoff_mteb_2023}
N.~Muennighoff, N.~Tazi, L.~Magne, and N.~Reimers.
\newblock {MTEB}: {Massive} {Text} {Embedding} {Benchmark}.
\newblock In A.~Vlachos and I.~Augenstein, editors, \emph{Proceedings of the 17th {Conference} of the {European} {Chapter} of the {Association} for {Computational} {Linguistics}}, pages 2014--2037, Dubrovnik, Croatia, May 2023. Association for Computational Linguistics.
\newblock \doi{10.18653/v1/2023.eacl-main.148}.
\newblock URL \url{https://aclanthology.org/2023.eacl-main.148}.

\bibitem[Lee et~al.(2024)Lee, Shakir, Koenig, and Lipp]{lee_open_2024}
S.~Lee, A.~Shakir, D.~Koenig, and J.~Lipp.
\newblock Open {Source} {Strikes} {Bread} - {New} {Fluffy} {Embeddings} {Model}, 2024.
\newblock URL \url{https://www.mixedbread.ai/blog/mxbai-embed-large-v1}.

\bibitem[Wang et~al.(2024)Wang, Yang, Huang, Yang, Majumder, and Wei]{wang_improving_2024}
L.~Wang, N.~Yang, X.~Huang, L.~Yang, R.~Majumder, and F.~Wei.
\newblock Improving {Text} {Embeddings} with {Large} {Language} {Models}, Jan. 2024.
\newblock URL \url{http://arxiv.org/abs/2401.00368}.
\newblock arXiv:2401.00368 [cs].

\bibitem[Levine et~al.(2020)Levine, Kumar, Tucker, and Fu]{levine_offline_2020}
S.~Levine, A.~Kumar, G.~Tucker, and J.~Fu.
\newblock Offline {Reinforcement} {Learning}: {Tutorial}, {Review}, and {Perspectives} on {Open} {Problems}, Nov. 2020.
\newblock URL \url{http://arxiv.org/abs/2005.01643}.
\newblock arXiv:2005.01643 [cs, stat].

\bibitem[Shankar et~al.(2021)Shankar, Elbaum, and Detweiler]{shankar2021freyja}
A.~Shankar, S.~Elbaum, and C.~Detweiler.
\newblock Freyja: A full multirotor system for agile \& precise outdoor flights.
\newblock In \emph{2021 IEEE International Conference on Robotics and Automation (ICRA)}, pages 217--223. IEEE, 2021.

\bibitem[Lillicrap et~al.(2016)Lillicrap, Hunt, Pritzel, Heess, Erez, Tassa, Silver, and Wierstra]{lillicrap_continuous_2016}
T.~P. Lillicrap, J.~J. Hunt, A.~Pritzel, N.~Heess, T.~Erez, Y.~Tassa, D.~Silver, and D.~Wierstra.
\newblock Continuous control with deep reinforcement learning.
\newblock In Y.~Bengio and Y.~LeCun, editors, \emph{4th {International} {Conference} on {Learning} {Representations}, {ICLR} 2016, {San} {Juan}, {Puerto} {Rico}, {May} 2-4, 2016, {Conference} {Track} {Proceedings}}, 2016.
\newblock URL \url{http://arxiv.org/abs/1509.02971}.

\end{thebibliography}

\appendix
\section{SARSA and Expected SARSA}
\label{sec:appendix_sarsa}

SARSA \citep{sutton_reinforcement_2018} is an on-policy method that learns the following Q function
\begin{align}
    Q_\pi(s, a) = R(s, a, s') + \gamma Q_\pi(s', a'),
\end{align}
where $a'$ is the recorded action that taken in the environment. Notice the difference compared to standard Max Q learning, which utilizes a max operator and potentially out-of-distribution action $a'$
\begin{align}
    Q(s, a) = R(s, a, s') + \gamma \max_{a' \in A} Q(s', a').
\end{align}

While Max Q learns the Q function for an optimal policy, SARSA learns a Q function for the policy that generated $a'$. Given these considerations, \emph{SARSA is not expected to converge to an optimal policy} like Max Q is. Only as $\pi$ approaches $\max_{a' \in A} Q_*(s, a)$ does expected SARSA approach the optimal policy. Although counter-intuitive, \citep{sutton_reinforcement_2018} shows how this suboptimal policy can be beneficial in the cliff walking scenario. In this scenario, the Max Q policy learns an optimal, but dangerous policy walking near the edge of a cliff to more quickly arrive at the goal. On the other hand, SARSA learns a safer policy that moves away from the cliff at the expense of a slightly reduced return. Even though Max Q should perform better in theory, in practice the Q function is imperfect and Max Q performs worse than SARSA \citep{sutton_reinforcement_2018}.

\begin{figure}[h]
    \centering
    \includegraphics[width=0.5\linewidth]{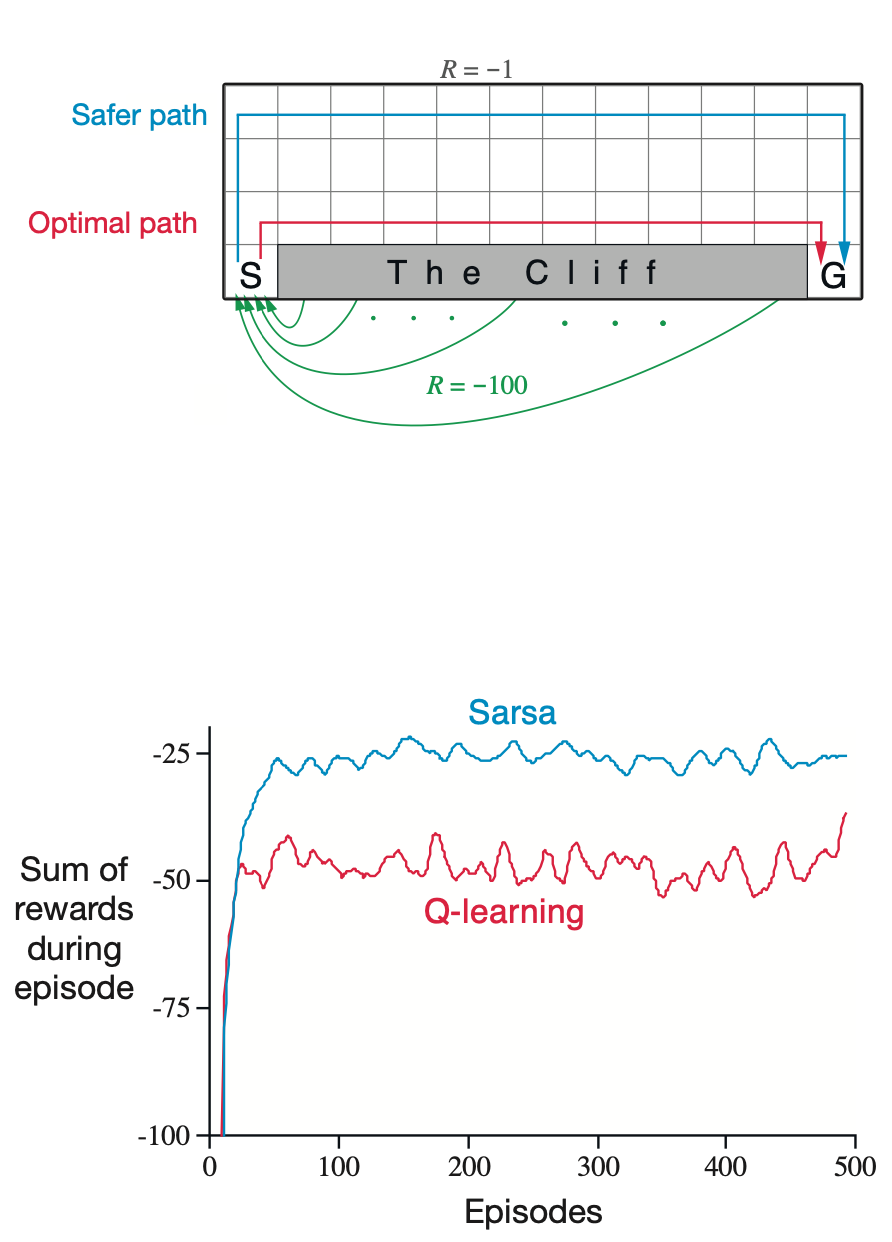}
    \caption{SARSA compared with Max Q, as it appears in \citep{sutton_reinforcement_2018}. SARSA learns a safer policy than Max Q on the cliff walking scenario. Max Q learns an optimal policy that remains close to the cliff edge, but reaches the goal more quickly. A single mistake can send the Max Q policy over the edge. In this case, SARSA samples uniform actions with probability $0.1$ and behaves greedily otherwise, making it beneficial to create a gap between the agent and the edge. This results in a safer but less optimal policy. Even though Max Q should produce more optimal policies in theory, SARSA can outperform Max Q in practice.}
    \label{fig:enter-label}
\end{figure}

While the SARSA update utilizes actions $a'$ from the dataset, Expected SARSA uses the policy action distribution instead
\begin{align}
    Q_\pi(s, a) = R(s, a, s') + \gamma \sum_{a' \in A} \pi(a | s) Q_\pi(s', a').
\end{align}

\subsection{Expected SARSA as a One Line Change to Q-Learning}
In the text, we allude to a one-line code change to obtain expected SARSA from Max Q. Therefore, we demonstrate this change with some pseudocode. Here, $q$ represents the Q function, $g$ is the decay factor $\gamma$, $s, a, r, sp$ correspond to the state, action, reward, and next state, $d$ corresponds to the episode termination flag, and $t$ corresponds to temperate $\tau$. The difference between Max Q and Expected SARSA is how we compute \texttt{next\_q\_value}.

\begin{lstlisting}[language=Python,escapeinside={(*}{*)}]
    def max_q(q, g, s, a, r, sp, d):
        q_value = q(s)[a]
        next_q_value = q(sp).max()
        target = r + !d * g * next_q
        return q_value - target
\end{lstlisting}

\begin{lstlisting}[language=Python,escapeinside={(*}{*)}]
    def mean_q(q, g, s, a, r, sp, d):
        q_value = q(s)[a]
        next_q_value = q(sp).mean()
        target = r + !d * g * next_q
        return q_value - target
\end{lstlisting}

\begin{lstlisting}[language=Python,escapeinside={(*}{*)}]
    def soft_q(q, g, s, a, r, sp, d, t):
        q_value = q(s)[a]
        next_q_value = (softmax(q(sp) / t) * q(sp)).sum()
        target = r + !d * g * next_q
        return q_value - target
\end{lstlisting}

\subsection{The Relationship Between Offline Expected SARSA and Importance Sampling}
We can view the offline Expected SARSA formulation as similar to Importance Weighted Off-Policy SARSA \citep[7.3]{sutton_reinforcement_2018}. If we assume a uniform prior $\pi_\beta$, then the importance sampling weight is given by
\begin{align}
    \frac{\pi(s)}{\pi_\beta(s)} = \frac{\pi(s)}{1 / |A|} = |A|\pi(s)
\end{align}
because
\begin{align}
     \pi_\beta(s) = \mathcal{U}(A), \pi_\beta(a | s) = \frac{1}{|A|} \quad \forall a, s.
\end{align}
Since $|A|$ is a constant, we can say that the Expected SARSA Q values are proportional to those learned by Off-Policy SARSA under importance sampling.

\section{Additional Experiments}
\label{sec:appendix_experiments}
We provide a number of additional experiments. We perform ablations on CQL and Soft Q hyperparameters, evaluate CQL and Max Q in the real world, and further inspect the results of our five-agent robot experiments. Finally, we briefly discuss a decentralized form of our method.
\subsection{CQL and Soft Q Ablation Studies}
\label{sec:reg_ablation}
\begin{figure}[h]
    \centering
    \includegraphics[width=0.48\linewidth]{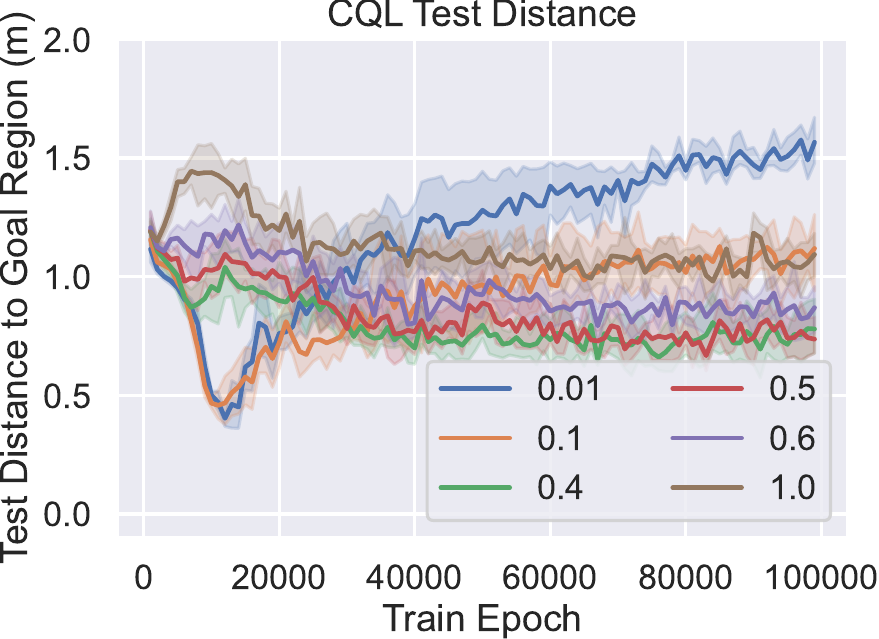}
    \includegraphics[width=0.48\linewidth]{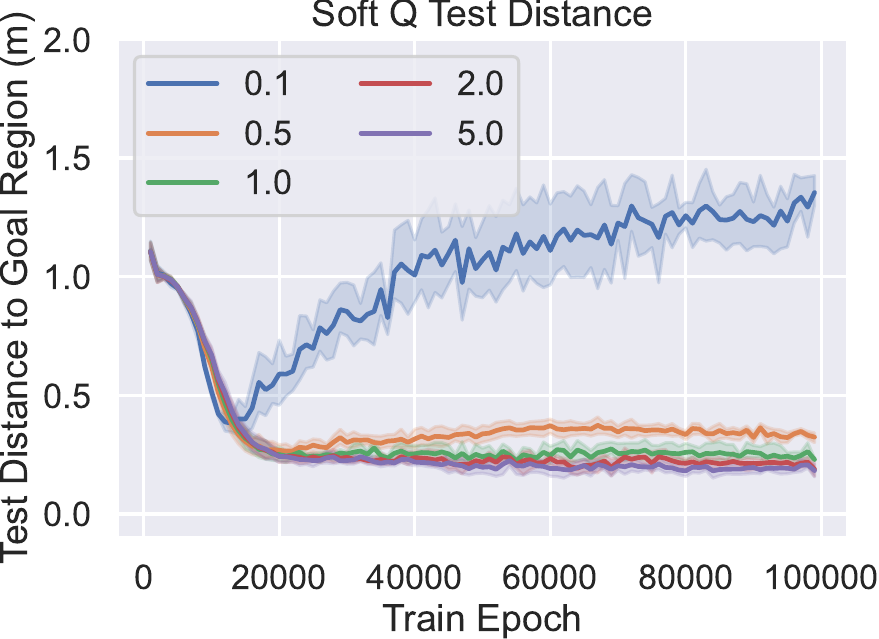}
    
    \includegraphics[width=0.48\linewidth]{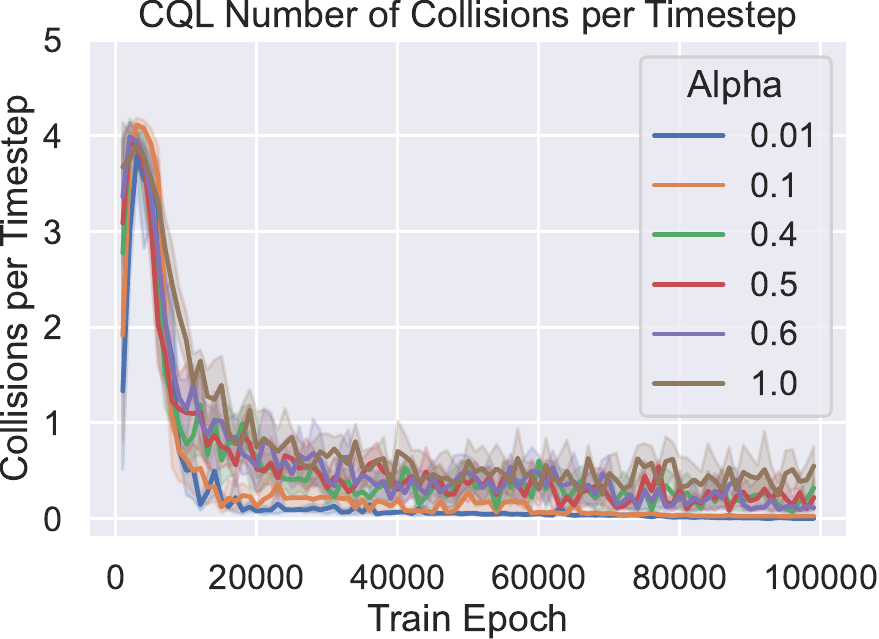}
    \includegraphics[width=0.48\linewidth]{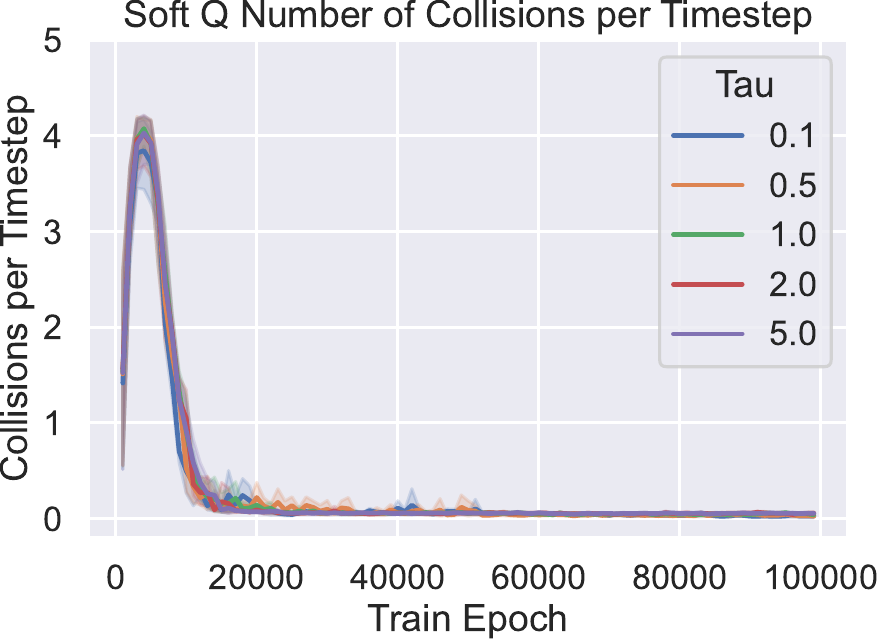}    
    \caption{Comparing CQL and Expected SARSA soft Q losses in simulation. (Top) We consider CQL with a variety of regularization strengths $\alpha$ and Expected SARSA Soft Q with a number of temperatures $\tau$. We inspect both the mean distance to the goal over an episode, as well as the number of collisions with the boundaries or other agents over an episode. Greater CQL regularization appears to reduce overextrapolation at the cost of a largely suboptimal policy. Expected SARSA Soft Q exhibits more stable training, resulting in more optimal policies with virtually no collisions. Note that as we decrease $\tau$, the Soft Q objective approaches Max Q, and we observe overextrapolation for $\tau = 0.1$ and to a lesser extent $\tau = 0.5$}
    \label{fig:reg_ablate}
\end{figure}

To find ideal CQL regularization strengths $\alpha$ and soft Q temperatures $\tau$, we perform an ablation in \cref{fig:reg_ablate}. We spawn the agents randomly in the arena, and sample random goals for each agent. We repeat this test five times every 1000 updates of the Q network.

\subsection{Max Q and CQL in the Real-World}
\label{sec:bad_real}
\begin{figure}[h]
    \centering
    \includegraphics[width=0.45\linewidth]{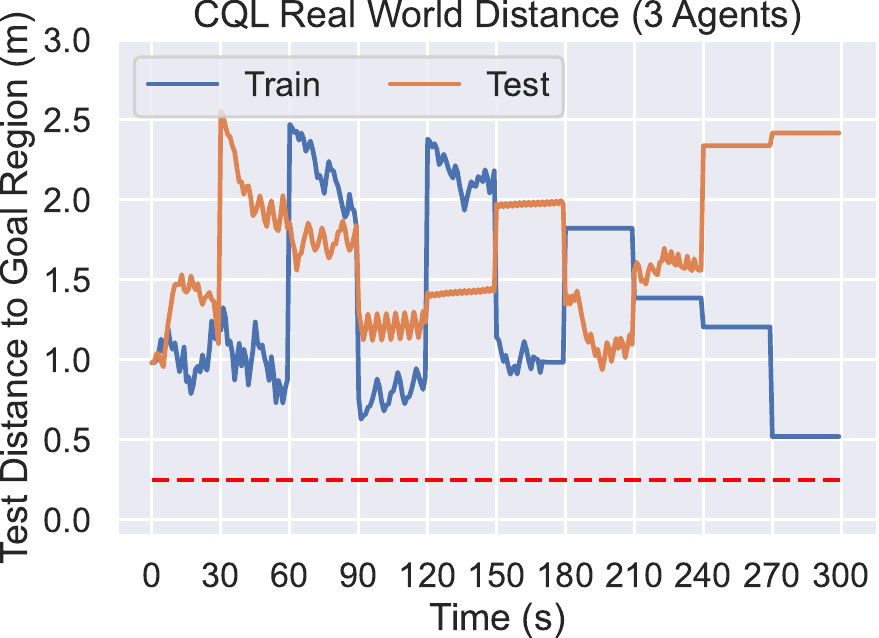}
    \includegraphics[width=0.45\linewidth]{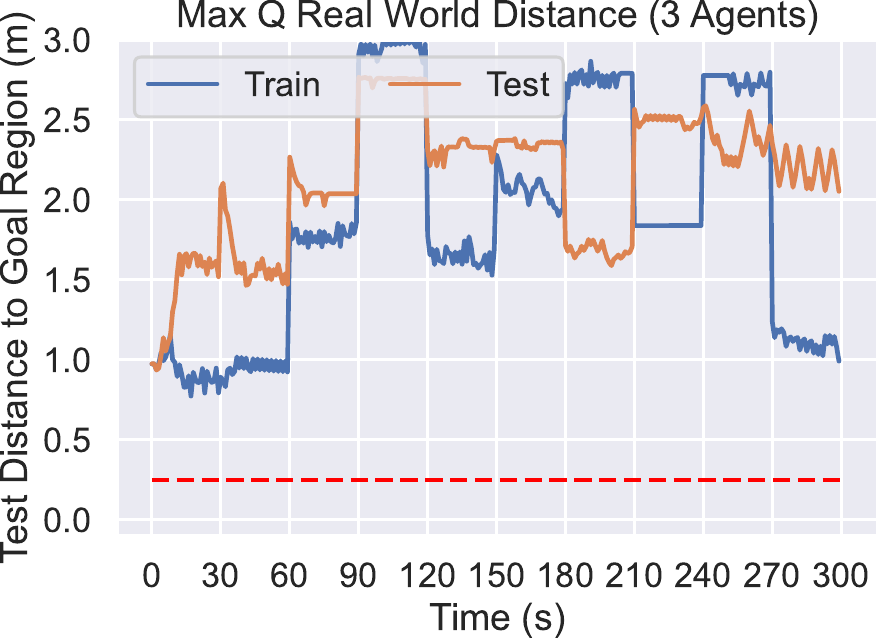}
    \caption{The real-world experiments for the CQL and Max Q objectives. We report the distance to goal, taking the mean across all agents. Neither method performs particularly well.}
    \label{fig:bad_real}
\end{figure}

\begin{figure}[h]
    \centering
    \includegraphics[width=0.45\linewidth]{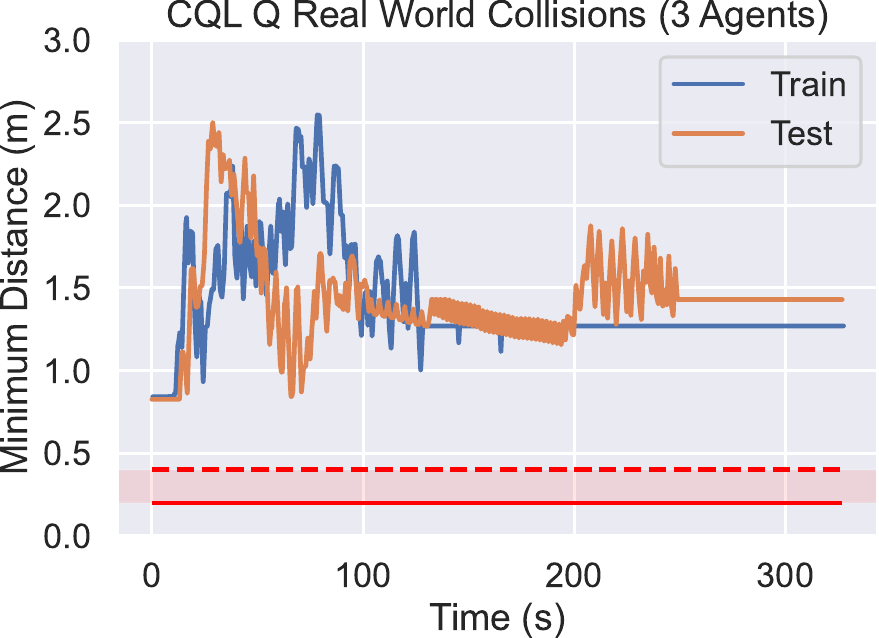}
    \includegraphics[width=0.45\linewidth]{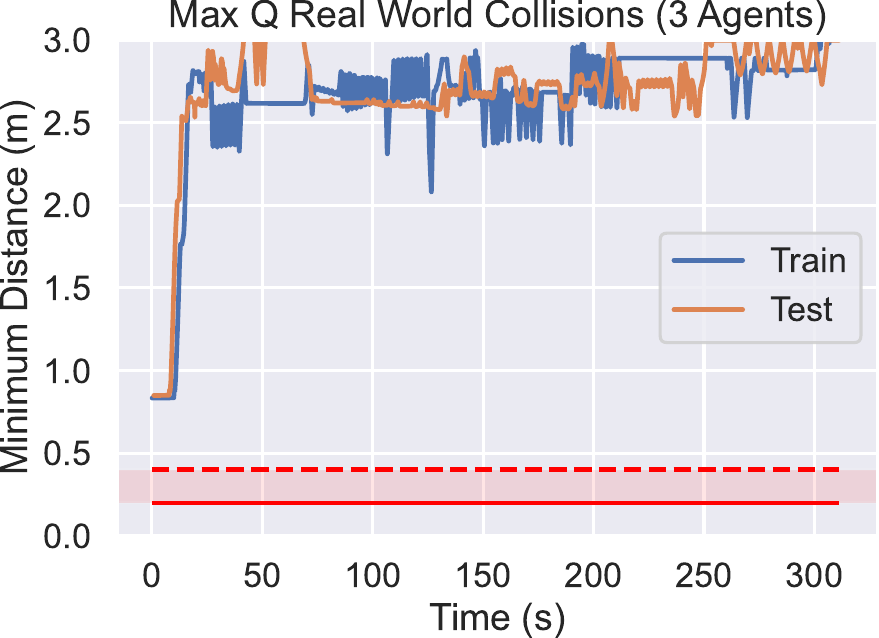}
    \caption{The real-world experiments for the CQL and Max Q objectives. We report the number of collisions, taking the mean across all agents. Both methods are capable of avoiding collisions and remaining within the boundaries. Any distances below the dotted red line are where agents breach their safety boundary of \SI{40}{cm}, while the solid red line denotes a collision in the physical world.}
    \label{fig:bad_real_collision}
\end{figure}

In this subsection, we provide results for the Max Q and CQL objectives in the real world. Similar to simulation, the agents learn to maximize the distance between themselves to avoid collisions, but never manage to learn to navigate to the provided goals (\cref{fig:bad_real}, \cref{fig:bad_real_collision}).

\subsection{Investigating Real-World Collisions}
\label{sec:real_collisions}
Although we did not observe any collisions in the three-agent real-world scenarios, we observed two minor grazes (less than a second of contact on diagonal points) in the five-agent real-world experiments. Let us further investigate these collisions. 

\begin{figure}
    \centering
    \begin{subfigure}{0.48\linewidth}
        \includegraphics[width=\linewidth]{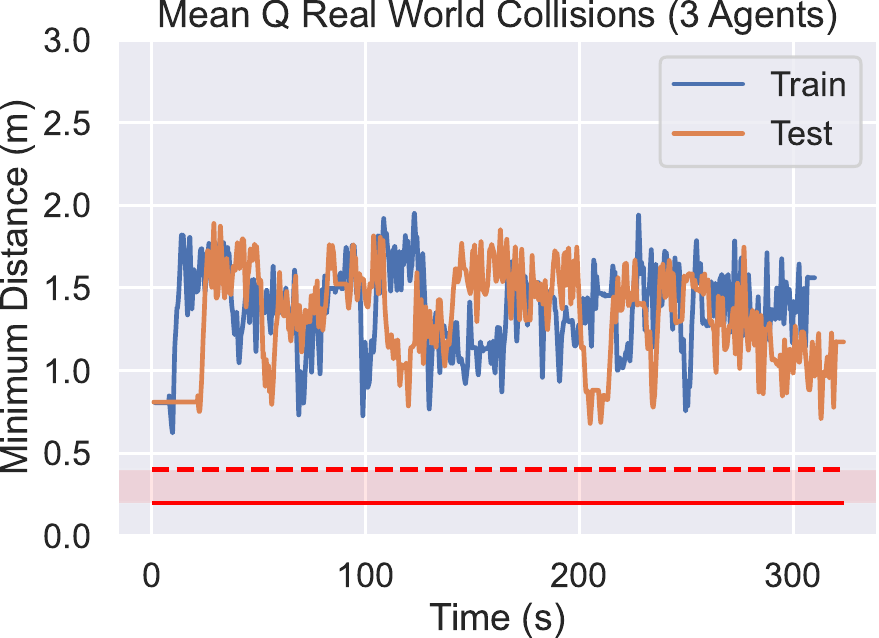}
    \end{subfigure} \hfill
    \begin{subfigure}{0.48\linewidth}
        \includegraphics[width=\linewidth]{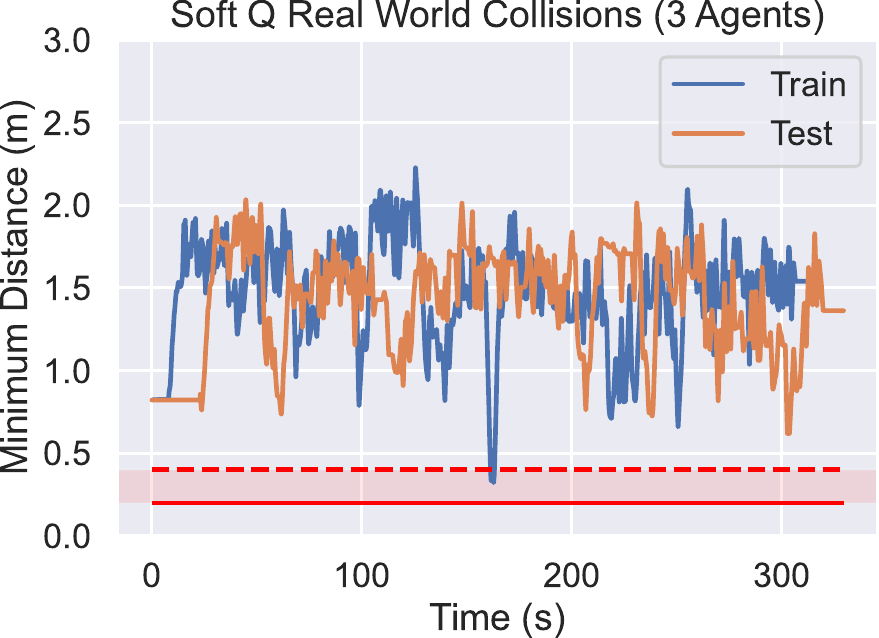}
    \end{subfigure}
    \begin{subfigure}[c]{0.48\linewidth}
        \includegraphics[width=\linewidth]{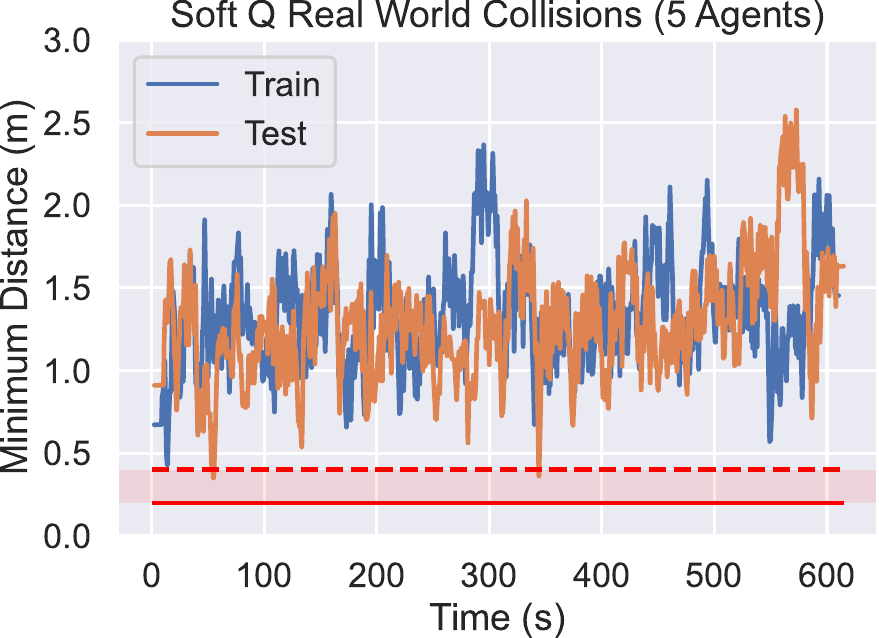}
    \end{subfigure}\hfill
    \begin{subfigure}[c]{0.48\linewidth}
        \includegraphics[width=\linewidth]{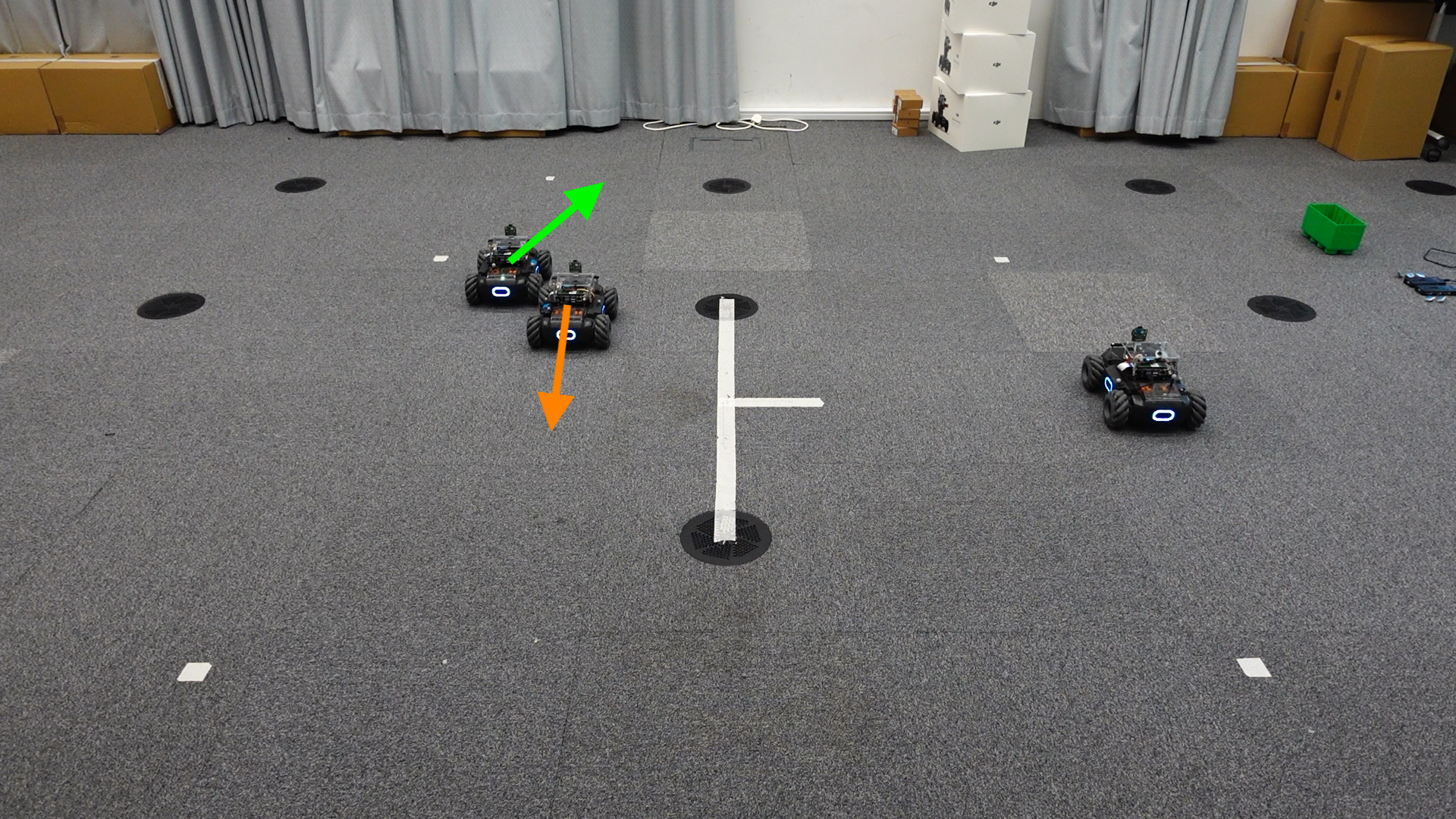}
    \end{subfigure}
    \caption{We plot the minimum distance observed between all robots, as reported by our motion capture system and recorded by ROS2. Any distances below the dotted red line are where agents breach their safety boundary of \SI{40}{cm}, while the solid red line denotes a collision in the physical world. Although we observe two minor scrapes in the five-agent videos, our data suggests that no collisions occurred. The image on the bottom right corresponds to point in time with the lowest minimum distance for the three-agent Soft Q objective (around 160s). Even though the safety boundary is breached, there is no physical collision.}
    \label{fig:real_collision}
\end{figure}

In our investigation, we visualise the positions of the robots, as logged by ROS and motion capture system in \cref{fig:real_collision}. 
Interestingly, we find that according to our raw logs, \textit{no agents} ever collide even though they might approach quite close to each other.
The shaded red region in the figure indicates our collision thresholds, and we find that the minimum distance between all robots is always above the hard threshold (solid red).

We attribute this to small offsets in the rigid-body's centroid (as tracked by motion capture) compared to the actual geometric pivot of the robots.
This is primarily why the logs don't show a collision.
These offsets produce compounded effects since the robots have a rectangular base (i.e., the diagonal may appear artificially shifted), and thus the controller's outputs can induce a slight twist on the body.
Furthermore, in our ROS ecosystem, we run motion capture in a best-effort mode, and observe rare spikes in network latency due to congestion (especially when high-frequency logging is active).
We note in \cref{fig:real_collision} (bottom-right) that our policies do not generate actions that lead to obvious collisions: the two robots approach very close, but their actions are pointed in different directions.
These close approaches may be avoided by increasing the distance threshold further.


\subsection{Distributed Policies}
Although we execute our policies in a centralized manner in our experiments, there also exists a decentralized equivalent. Using decentralized policies enables agents to compute and take actions on their own. Let us rewrite our embeddings $z_i$, states $s_i$, and actions $a_i$ to make this clear
\begin{align}
    z_i &= \phi(g_i) \\
    s_i &= f(o_i, z_i, o_{1:(i-1), (i+1):n}, z_{1:(i-1), (i+1):n}) \\
    \pi(a \vert s_i) &= \frac{e^{Q(s_i, a) / \tau}}{\sum_{\alpha \in A} e^{Q(s_i, \alpha) / \tau}} \\
    a_i &\sim \pi(a \vert s_i) 
\end{align}
We see that each agent's action $a_i$ depends only on the observations $o$ and latent tasks $z$ of itself and the other agents. Once we have computed the state $s_i$, the policies can independently compute the action $a_i$. In other words, all computation ($f, \pi$) can be computed independently.

\clearpage

\section{Experiment Details}
In this section, we provide further details for the LLM latent space experiment and the RL experiments.

\subsection{LLM Latent Space Details}
For the LLM decoder experiment \cref{fig:llm-compare}, we utilize a $k = \SI{1.1}{m}$ and train using mean squared error. Our neural network consists of three blocks $B$ (linear layer, layernorm, and leaky ReLU, see \cref{sec:training}) followed by a final linear layer.

\begin{table}[h]
    \centering
    \begin{tabular}{lr}
        Hyperparameter & Value \\
        \hline
         Batch size & 32\\
         Learning rate & 0.00001\\
         Learning rate warmup epochs & 0\\
         Optimizer & Adamw\\
         Weight decay & 0.0001\\
         $|z|$ & 768\\
         Hidden size & 256\\
    \end{tabular}
    \vspace{1em}
    \caption{Decoder hyperparameters for the LLM analysis experiment.}
    \label{tab:decoder_hparam}
\end{table}

\subsection{Real-World Experimental Setup and Framework}
\label{sec:appendix_exp_setup}
Here we provide additional details on the robots as well as the framework we use for all our experiments.
The robotic platforms we use are DJI Robomaster S1 platforms, which are customized similar to prior work~\cite{blumenkamp_cambridge_2024}.
The robots have an omnidirectional mecanum-wheel drive, and can track fixed-frame position and velocity state references using an onboard control stack (\textit{Freyja}~\cite{shankar2021freyja}). The robots have a radius of approximately \SI{25}{cm}. We operate within a \qtyproduct{3.8 x 3.8}{m} subset of a larger \qtyproduct{6 x 4}{m} motion-capture arena.
We use the default North-East-Down frame convention (with the Down component always zero for ground robots), which is why some of the text prompts contain cardinal directions (\texttt{`north edge'}) in them.

The actions generated by our trained policies are fixed-frame velocity targets for the robot, commanded at \SI{1}{Hz}, and are tracked by the onboard controller at \SI{50}{Hz}.
Note that since our policies take $<\SI{2}{ms}$ to generate actions, we can potentially run them at substantially higher rates if necessary.
However, with discretized actions, we find that an action frequency of \SI{1}{Hz} is sufficient for the navigation tasks we consider.
Our control stack, motion-capture setup and the policy execution are all wrapped in a ROS2 ecosystem, and the exact framework is also utilized for collecting real-world datasets.

\subsection{RL Training Details}
\label{sec:appendix_train_details}
We list the hyperparameters used for both the simulated and real-world RL experiments in \cref{tab:hparam}.
\begin{table}[h]
    \centering
    \begin{tabular}{lr}
        Hyperparameter & Value \\
        \hline
         Decay $\gamma$ & 0.95\\
         Polyak $\tau$ & 0.0005\\
         Batch size & 256\\
         Learning rate & 0.0001\\
         Learning rate warmup & linear\\
         Learning rate warmup epochs & 1000\\
         Optimizer & Adamw\\
         Weight decay & 0.0001\\
         $|z|$ & 768\\
         Hidden size & 1024\\
         LLM & GTE-Base\\
         Ensemble size & 2\\
         Ensemble reduce & $\min$\\
         \multicolumn{2}{c}{Simulation}\\
         \hline
         Evaluation episodes per epoch & 5\\
         Timesteps per evaluation episode & 50\\
         Collision radius & \SI{40}{cm}\\
         Epochs trained  & 100k\\
         \multicolumn{2}{c}{Real World}\\
         \hline
         Evaluation episodes & 10 or 20\\
         Timesteps per evaluation episode & 30\\
         Epochs trained & 1M \\
         Collision radius & \SI{20}{cm} - \SI{30}{cm}\\
\end{tabular}
    \vspace{1em}
    \caption{Hyperparameters shared between all RL experiments.}
    \label{tab:hparam}
\end{table}
We provide an arena to the agents, outside of which is considered a collision. During training, we approximate the collision radius of each agent as \SI{0.4}{m}. During deployment, our actions come from a Boltzmann policy with a temperature of $\tau = 0.01$
\begin{align}
    \pi(a | s_i) &= \frac{e^{Q(s, a) / \tau}}{\sum_{\alpha \in A} e^{Q(s, \alpha) / \tau}} \\
    a_i &\sim \pi(a | s_i).
\end{align}
Each agent $i$ shares the same parameters for $\pi$. We update the Q function using a Polyak soft update \citep{lillicrap_continuous_2016}.

\clearpage
\section{Tasks}
\label{sec:appendix_tasks}
We provide a list of commands and locations that we use to generate tasks. All tasks are in the form \texttt{Agent, <command to location>}. We generate train commands by combining a location with a command string. The \texttt{\{\}} denote where the location goes in the command string.

\begin{minipage}{0.45\linewidth}
\begin{verbatim}
navigate to the {}
pathfind to the {}
find your way to the {}
move to the {}
your goal is the {}
make your way to the {}
head towards the {}
travel to the {}
reach the {}
proceed to the {}
go to the {}
the {} is your target
\end{verbatim}
\end{minipage}\hfill\begin{minipage}{0.45\linewidth}
\begin{verbatim}
west edge
east edge
south edge
north edge
left edge
right edge
bottom edge
top edge
lower edge
upper edge
south west corner
west south corner
south east corner
east south corner
north west corner
west north corner
north east corner
SW corner
SE corner
NW corner
NE corner
bottom left corner
left bottom corner
bottom right corner
right bottom corner
top left corner
left top corner
top right corner
right top corner
lower left corner
left lower corner
lower right corner
right lower corner
upper left corner
left upper corner
upper right corner
right upper corner
\end{verbatim}
\end{minipage}

\clearpage

\end{document}